%% file: main.tex
\pdfoutput=1
 \documentclass[conference, 10pt]{IEEEtran}

\IEEEoverridecommandlockouts                              

\usepackage{graphics} 
\usepackage{epsfig} 
\usepackage{mathptmx} 
\usepackage{mathrsfs}
\usepackage{times} 
\usepackage{amsmath} 

\usepackage{amsthm}
\usepackage{amssymb}  
\usepackage{wasysym}
\usepackage{txfonts}
\usepackage{bbm}
\usepackage[singlelinecheck=false,font=footnotesize]{caption}
\usepackage{subcaption}
\usepackage{color}  
\usepackage[dvipsnames]{xcolor}
\usepackage{indentfirst}
\usepackage{multirow}

\usepackage{perl_acronyms}
\usepackage{perl_misc}

\usepackage{color, colortbl}
\usepackage{algorithm}
\usepackage{algpseudocode}

\usepackage{geometry}
\usepackage[
    style=ieee,
    doi=false,
    isbn=false,
    url=false,
    eprint=false,
    backend=bibtex,
    natbib=true
    ]{biblatex}

 \usepackage{nopageno}

\bibliography{references}

\usepackage{armtd_commands}

\usepackage{hyperref}

\title{\LARGE \bf Reachable Sets for Safe, Real-Time Manipulator Trajectory Design}
\author{Patrick Holmes$^1$, Shreyas Kousik$^1$, Bohao Zhang$^1$, Daphna Raz$^2$,\\
Corina Barbalata$^3$, Matthew Johnson-Roberson$^{2,4}$, Ram Vasudevan$^{1,2}$
\thanks{This work is supported by the Ford Motor Company via the Ford-UM Alliance under award N022977, and the Office of Naval Research under award number N00014-18-1-2575.}
\thanks{$^{1}$Mechanical Engineering, University of Michigan, Ann Arbor, MI. {\tt\small <pdholmes, skousik, jimzhang, ramv>@umich.edu}}%
\thanks{$^{2}$Robotics Institute, University of Michigan, Ann Arbor, MI. {\tt\small daphraz@umich.edu}}%
\thanks{$^{3}$Mechanical and Industrial Engineering, Louisiana State University, Baton Rouge, LA. {\tt\small cbarbalata@lsu.edu}}%
\thanks{$^{4}$Naval Architecture and Marine Engineering, University of Michigan, Ann Arbor, MI. {\tt\small mattjr@umich.edu}}%
}
\begin{document}

\setlength{\textfloatsep}{18pt}
\maketitle
\thispagestyle{empty}
\pagestyle{plain}

\begin{abstract}
For robotic arms to operate in arbitrary environments, especially near people, it is critical to certify the safety of their motion planning algorithms.
However, there is often a trade-off between safety and real-time performance; one can either carefully design safe plans, or rapidly generate potentially-unsafe plans.
This work presents a receding-horizon, real-time trajectory planner with safety guarantees, called ARMTD (Autonomous Reachability-based Manipulator Trajectory Design).
The method first computes (offline) a reachable set of parameterized trajectories for each joint of an arm.
Each trajectory includes a fail-safe maneuver (braking to a stop).
At runtime, in each receding-horizon planning iteration, ARMTD constructs a parameterized reachable set of the full arm in workspace and intersects it with obstacles to generate sub-differentiable, provably-conservative collision-avoidance constraints on the trajectory parameters.
ARMTD then performs trajectory optimization over the parameters, subject to these constraints.
On a 6 degree-of-freedom arm, ARMTD outperforms CHOMP in simulation, never crashes, and completes a variety of real-time planning tasks on hardware.
\end{abstract}

\input{sections/01_introduction.tex}
\input{sections/02_arm_representation.tex}

\input{sections/03_reachability.tex}
\input{sections/04_online_planning.tex}

\input{sections/05_demos.tex}
\input{sections/06_conclusion.tex}




\renewcommand{\bibfont}{\normalfont\small}
{\renewcommand{\markboth}[2]{}
\printbibliography}

\clearpage

\begin{appendices}
\input{sections/appendix_supplement}
\end{appendices}

\end{document}

%% file: sections/01_introduction.tex
\section{Introduction}\label{sec:intro}

\begin{figure}[t]
    \centering
    \includegraphics[width=0.95\columnwidth]{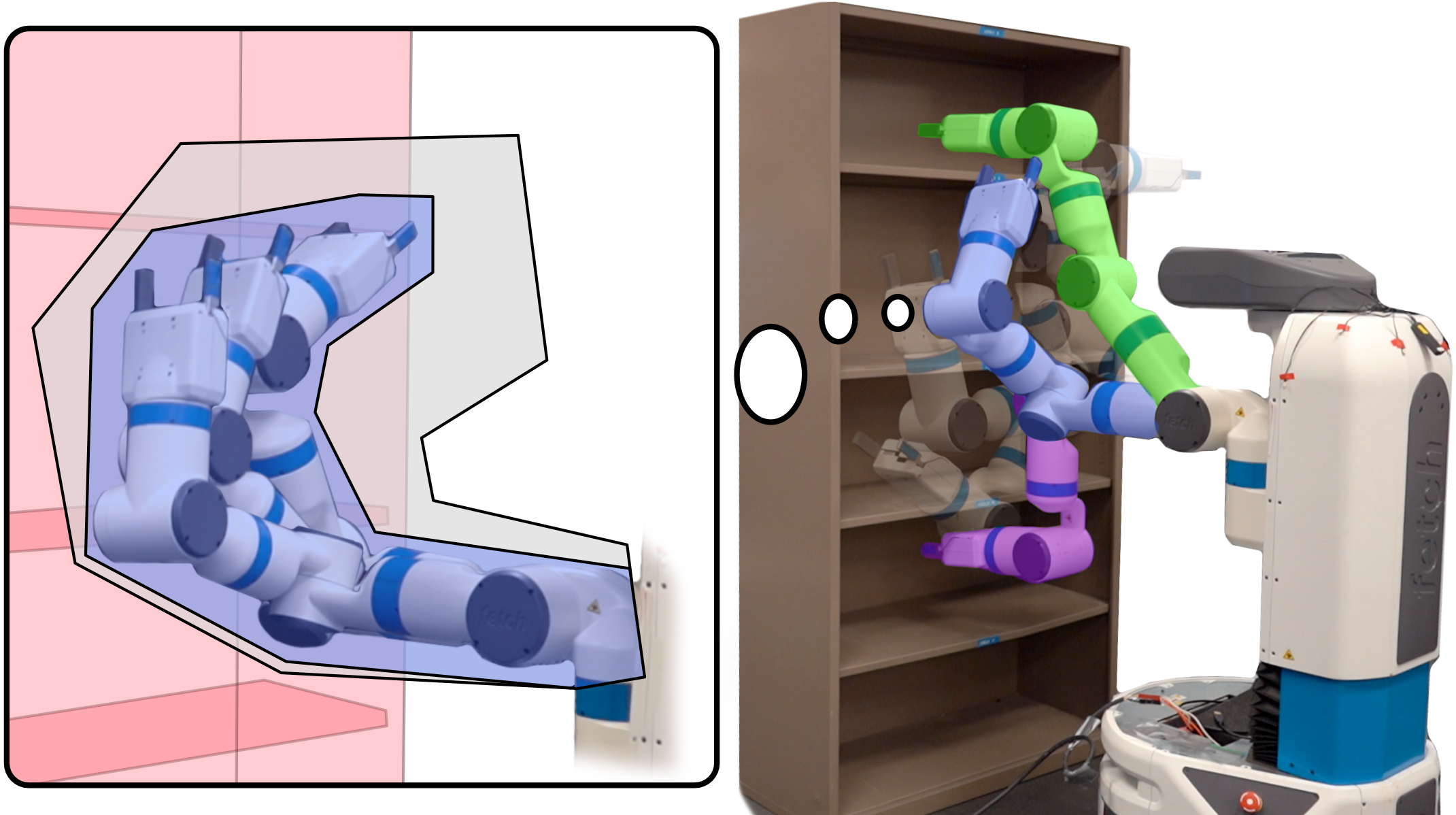}
    \caption{ARMTD performs safe, real-time receding-horizon planning for a Fetch arm around a cabinet in real time, from a start pose (purple, low shelf) to a goal (green, high shelf).
    Several intermediate poses are shown (transparent).
    The callout on the left, corresponding to the blue intermediate pose, shows a single planning iteration, with the shelf in light red.
    In grey is the arm's reachable set for a continuum of parameterized trajectories over a short time horizon.
    The smaller blue set is the subset of the reachable set corresponding to the particular trajectory that was selected for this planning iteration, which is guaranteed not to collide with the obstacle.
    Over many such trials in simulation and on hardware, ARMTD never crashed.
    See our video: \href{https://youtu.be/ySnux2owlAA}{\texttt{youtu.be/ySnux2owlAA}}.}
    \label{fig:fetch_intro}
    \vspace*{-0.5cm}
\end{figure}

To maximize utility in arbitrary environments, especially when operating near people, robotic arms should plan collision-free motions in real time.
Such performance requires sensing and reacting to the environment as the robot plans and executes motions; in other words, it must perform \defemph{receding-horizon planning}, where it iteratively generates a plan while executing a previous plan.
This paper addresses \textbf{guaranteed-safe receding-horizon trajectory planning for robotic arms}.
We call the proposed method Autonomous Reachability-based Manipulator Trajectory Design, or \textbf{ARMTD}, introduced in Fig. \ref{fig:fetch_intro}.

Motion planning can be broadly split into three paradigms, depending on whether safety is enforced by (1) a path planner, (2) a trajectory planner, or (3) a tracking controller.

The first paradigm is commonly used for robotic arm planning, wherein the path planner is responsible for safety.
One generates a collision-free path, then smooths it and parameterizes it by time (i.e., converts it into a trajectory) \cite{pfeiffer1987_old_traj_planning,Kunz2012_traj_gen}.
Such methods often have a tradeoff between safety and real-time performance because they represent paths with discrete points in configuration space  \cite{lavalle2001randomized,kavraki1996probabilistic}.
Ensuring safety requires approximations such as buffering the volume of the arm at each discrete point to account for the discretization, or computing the swept volume along the path assuming, e.g., straight lines between points \cite{lavalle_textbook}.
If one treats the path as a decision variable in a nonlinear optimization program, the gradient of the distance between the arm's volume and obstacles may ``push'' each configuration out of collision \cite{chomp,trajopt,itomp}.
This means the output path can be treated directly as a trajectory, if the optimization uses path smoothness as the cost.
However, this relies on several approximations to achieve real-time performance: finite differencing to bound joint speeds and accelerations, collision penalties in the cost instead of hard constraints, and finite differencing \cite{chomp} or linearization \cite{trajopt} for the collision-avoidance penalty gradient.
This necessitates finer discretization to faithfully represent the robot's kinematics.
To enable real-time performance without gradients, one can compute many paths offline, then collision-check at runtime \cite{Murray2016_motion_planning_on_chip,kunz2010_PRM}; but for arbitrary tasks, it can be unclear how many paths are necessary, or how to ensure safety if the arm's volume changes (e.g., by grasping an object).
Another approach to real-time performance is to plan iteratively in a receding-horizon either by gradient descent (with the same drawbacks as above) \cite{itomp} or assuming the underlying path planner is safe \cite{Hauser2012_receding_horizon}.
In summary, in this paradigm, one must discretize finely, or buffer by a large amount, to achieve safety at the expense of performance.

In the second paradigm, the path planner generates a (potentially unsafe) path, then the trajectory planner attempts to track the path as closely as possible while maintaining safety.
In this paradigm, one computes a \defemph{reachable set} (RS) for a family of trajectories instead of computing a swept volume for a path.
Methods in this paradigm can achieve both safety and real-time performance in receding-horizon planning by leveraging sums-of-squares programming \cite{majumdar2016funnel,kousik2018_RTD_ijrr,Vaskov2019_RSS} or zonotope reachability analysis \cite{kousik2019_quadrotor_rtd}.
Unfortunately, the methods in this paradigm suffer from the curse of dimensionality, preventing their use with the high-dimensional models of typical arms.

In the third paradigm, one attempts to ensure safety via the tracking controller, instead of in a path or trajectory.
Here, one builds a supervisory safety controller for pre-specified trajectories \cite{althoff2019_safe_modular_bots} or a set of safe states \cite{singletary2019_arm_barrier_function}.
Another approach is to compute a safety buffer and associated controller using Hamilton-Jacobi reachability analysis \cite{herbert2017_fastrack,chen2018_hjb_decomposition}, but the curse of dimensionality has prevented applying this to arms.

To the best of our knowledge, RSs in manipulator planning have only been used for either collision-checking a single, precomputed trajectory \cite{althoff2019_safe_modular_bots, fraichard2012guaranteeing}, or for controlling to a predefined setpoint \cite{majumdar2014control}.
In contrast, our proposed ARMTD method generates RSs for a continuum of trajectories, allowing optimization over sets of safe trajectories.
Computing such RSs directly is challenging because of the high-dimensional configuration space and nonlinear transformation to workspace used for a typical arm \cite{kousik2018_RTD_ijrr,chen2018_hjb_decomposition}.

Our proposed ARMTD method overcomes these challenges by composing a high-dimensional RS in workspace from low-dimensional reachable sets of joint configurations.
ARMTD extends the second planning paradigm above by using these RSs to plan safe trajectories in real time.
The RS also provides subdifferentiable collision-avoidance, self-intersection, and joint limit constraints for trajectory optimization.
Importantly, the RS composition, constraint generation, and gradient evaluation are all parallelizable.

We now provide an overview of ARMTD, also shown in Fig. \ref{fig:method_overview}.
ARMTD begins by specifying a parameterized continuum of kinematic configuration space trajectories, each of which includes a fail-safe maneuver.
Offline, ARMTD computes parameterized joint reachable sets, or JRSs, of these trajectories in configuration space.
At runtime (in each receding-horizon), it constructs a parameterized RS from the precomputed JRSs.
ARMTD intersects the RS with obstacles to generate provably-correct safety constraints.
ARMTD then performs trajectory optimization over the parameters, subject to the safety constraints.
If it cannot find a feasible solution within a prespecified time limit, the arm continues executing the trajectory from its previous planning iteration (which includes a fail-safe maneuver), guaranteeing perpetual safety \cite{kousik2018_RTD_ijrr, Hauser2012_receding_horizon}.
In this work, we only discuss static environments, but this approach can extend to dynamic environments \cite{Vaskov2019_RSS}.


\subsection{Contributions}
We make the following contributions.
First, a method to conservatively construct the RS of high-dimensional redundant robotic manipulators (Sections \ref{sec:reachability}--\ref{sec:online_planning}).
Second, a parallelized method to perform real-time, provably-safe, receding-horizon trajectory optimization (Section \ref{sec:online_planning}).
Third, a demonstration in simulation and on hardware, with no collisions (Section \ref{sec:demos} and Supplemental Video), plus a comparison to CHOMP \cite{chomp}.
The remaining sections are Section \ref{sec:preliminaries} (Arm, Obstacles, and Trajectory Parameters) and Section \ref{sec:conclusion} (Conclusion).
See our video: \href{https://youtu.be/ySnux2owlAA}{\texttt{youtu.be/ySnux2owlAA}}.
Our code is available: \href{https://github.com/ramvasudevan/arm_planning}{\texttt{github.com/ramvasudevan/arm\_planning}}.
All proofs, plus additional explanations, are available in the appendices included at the end of this document.



\subsection{Notation}
The $n$-dimensional real numbers are $\R^n$, natural numbers are $\N$, the unit circle is $\s^1$, and the set of $3\times 3$ rotation matrices is $\SO(3)$.
Vectors are either $[x_1,\cdots,x_n]^\top$ or $(x_1,\cdots,x_n)$ depending on if the size/shape is relevant.
Let $U, V \subset \R^n$.
For a point $p \in U$, $\{p\} \subset U$ is the set containing $p$.
The power set of $U$ is $\P(U)$.
The Minkowski sum is $U \oplus V = \{u + v~|~u \in U,\ v \in V\}$.
For a matrix $A \in \R^{n\times n}$, $AU = \{Au~|~u \in U\}$.
For matrices, $\prod$ performs right multiplication with increasing index (e.g., $\prod_{i=1}^3 A_i = A_1A_2A_3$).
Greek lowercase letters in angle brackets are indeterminate variables (e.g., $\sym{\sg}$).
Superscripts on points index elements of a set.
Subscripts are joint indices or contextual information.

%% file: sections/02_arm_representation.tex
\section{Arm, Obstacles, and Trajectory Parameters}\label{sec:preliminaries}

The goal of this work is to plan collision-free trajectories for a robotic arm operating around obstacles in a receding-horizon framework.
We now discuss the arm and its environment, then our receding-horizon framework and parameterized trajectories.

\subsection{Arm and Obstacles}

\subsubsection{Arm}
Consider an arm with $n_q \in \N$ joints (i.e., $n_q$ DOFs) and $n_q + 1$ links, including the $0$\ts{th} link, or \defemph{baselink}.
We make the following assumptions/definitions.
Each joint is a single-axis revolute joint, attached between a \defemph{predecessor} link and a \defemph{successor} link.
The arm is a single kinematic chain from baselink to end effector; link $i-1$ is joined to link $i$ by joint $i$ for $i = 1,\cdots,n_q$.
One can create multi-DOF joints using virtual links of zero volume.
The \defemph{configuration space} is $Q \subseteq \s^{n_q}$, containing \defemph{configurations} $q = (q_1, q_2, \cdots, q_{n_q}) \in Q$.
The space of joint velocities is $\dot{Q} \subset \R^{n_q}$.
There exists a default configuration $0 \in Q$.
The \defemph{workspace}, $W \subset \R^3$, is the all points in space reachable by any point on the arm in any configuration.
The robot's physical limits are as follows.
Each joint $i$ has a minimum and maximum position $\qlim^-$ and $\qlim^+$, maximum absolute speed $\dqlim$ and maximum absolute acceleration $\ddqlim$.

We now describe the kinematic chain.
Each link has a local coordinate frame with the origin located at the link's predecessor joint (the baselink's frame is the global frame).
The rotation matrix $R_i(q_i) \in \regtext{SO}(3)$ describes the rotation of link $i$ relative to link $i-1$ (by joint $i$).
The displacement $l_i \in \R^3$ denotes the position of joint $i$ on link $i$ relative to joint $(i-1)$ in the frame of link $i$.
The set $L_i \subset \R^3$ denotes the volume occupied by the $i$\ts{th} link, with respect to its predecessor joint, in the frame of link $i$.
Let $\FO_i: Q \to \P(W)$ give the forward occupancy of link $i$.
That is, the $i$\ts{th} link occupies the volume
\begin{equation}\label{eq:forward_occupancy_i}
    \FO_i(q)~=~\left\{\sum_{j < i}\Bigg(\prod_{n \leq j}R_n(q_n)\,l_j\Bigg)\right\} \oplus \left(\prod_{n\leq i}R_n(q_n) L_i \right) \subset W.
\end{equation}
Let $\FO: Q \to \P(W)$ give the occupancy of the entire arm: $\FO(q) = \bigcup_{i = 1}^{n_q} \FO_i(q)$.
\noindent Note, the first expression in \eqref{eq:forward_occupancy_i} gives the position of joint $(i-1)$ and the second gives the rotated volume of link $i$.
See Appendix \ref{apdx:explanation:FO} for an example.

\subsubsection{Obstacles}
We denote an \defemph{obstacle} as a set $O \subset W$.
If the arm's volume at $q \in Q$ is intersecting the obstacle, we say the arm is in \defemph{collision}, i.e. $\FO(q) \cap O \neq \emptyset$.
We assume the following about obstacles.
Each obstacle is compact and static with respect to time (note, one can extend ARMTD to dynamic obstacles \cite{Vaskov2019_RSS}).
At any time, there are at most $n\obs \in \N$, $(n\obs < \infty)$ obstacles in the workspace, and the arm has access to a conservative estimate of the size and location of all such obstacles (we are only concerned with planning, not perception).
Let $\obsset = \{O_1,\cdots,O_{n_O}\}$ denote a set of obstacles.


\subsection{Receding-Horizon Planning and Timing}
ARMTD plans in a receding-horizon way, meaning it generates a short plan, then executes it while generating its next short plan.
Every such plan is specified over a compact time interval $T \subset \R$.
Without loss of generality (WLOG), since time can be shifted to $0$ at the beginning of any plan, we denote $T = [0, \tfin]$.
We further specify that ARMTD must generate a new plan every $t\plan < \tfin$ seconds.
If a collision-free plan cannot be found within $t\plan$ s, the robot must continue the plan from the previous receding-horizon iteration; therefore, we include a fail-safe (braking) maneuver in each plan.
The durations $\tfin$ and $t\plan$ are chosen such that $(\tfin - t\plan)$ is large enough for the arm to stop from its maximum joint speeds given its maximum accelerations.
This ensures every plan can include a fail-safe maneuver. 
We abuse notation to let $q: T \to Q$ denote a trajectory plan and $q_i: T \to Q$ denote the trajectory of the $i^{th}$ joint.
A plan is \defemph{collision-free} if $\FO(q(t)) \cap O = \emptyset \, \forall t \in T,\ \forall\ O \in \obsset$.
Next, we specify the form of each plan.

\subsection{Trajectory Parameterization}\label{subsec:reachability_theory}

ARMTD plans using parameterized trajectories.
We describe the theory, then present our implementation.

\subsubsection{Theory}

Let $K \subset \R^{n_k}$, $n_k \in \N$, be a compact space of \defemph{trajectory parameters}, meaning each $k \in K$ maps to a trajectory $q: T \to Q$.
We use $q(t;k)$ to denote the configuration parameterized by $k \in K$ at time $t \in T$.
So, in each receding-horizon planning iteration, ARMTD attempts to select a single $k \in K$ (via trajectory optimization with obstacles represented as constraints on $K$.

\begin{defn}\label{def:traj_param_generic}
We require $q: T \to Q$ to satisfy three properties for all $k \in K$.
First, $q(\cdot\,; k)$ is at least once-differentiable w.r.t. time.
Second, $q(0;k) = 0$.
Third, $\dot{q}(\tfin;k) = 0$.
\end{defn}

\noindent 
The second property uses the fact that all joints are revolute, so $q(0 ;k) = 0$ WLOG.
The third property guarantees each parameterized trajectory includes a fail-safe braking maneuver.


Note, the parameterized trajectories are kinematic, not dynamic.
This is common in motion planning \cite{chomp,trajopt,itomp,Murray2016_motion_planning_on_chip,kunz2010_PRM}, because existing controllers can track such trajectories closely (e.g., within 0.01 rad for revolute joints \cite{paden1988globally,giusti2017_arm_traj_dyn_zono}) in the absence of disturbances such as collisions.
We find these trajectories sufficient to avoid collision in real-world hardware demonstrations (Sec. \ref{sec:demos}).
Also, methods exist for quantifying tracking error \cite{giusti2017_arm_traj_dyn_zono,kousik2019_quadrotor_rtd} and accounting for it at runtime \cite{kousik2018_RTD_ijrr,Vaskov2019_RSS}.

\subsubsection{Implementation}

We choose a parameterization that is simple yet sufficient for safe planning in arbitrary scenarios (see Sec. \ref{sec:demos}).
We define a \defemph{velocity parameter} $\kv \in \R^{n_q}$ for the initial velocity $\initdq$, and an \defemph{acceleration parameter} $\ka \in \R^{n_q}$ that specifies a constant acceleration over $[0, t\plan)$. 
We write $\kv = (\kv_1,\cdots,\kv_{n_q})$ and similarly for $\ka$.
We denote $k = (\kv,\ka) \in K \subset \R^{n_k}$, where $n_k = 2n_q$.
The trajectories are given by
\begin{align}
\label{eq:traj_parameterization}
    \dot{q}(t;k) = \begin{cases}
        \kv + \ka t, &t \in [0,t\plan) \\
        \frac{\kv + \ka t\plan}{\tfin - t\plan}(\tfin - t), &t \in [t\plan,\tfin],
    \end{cases},
\end{align}
with $q_i(0;k) = 0$ for all $k$ to satisfy Def. \ref{def:traj_param_generic}.
These trajectories brake to a stop over $[t\plan, \tfin]$ with constant acceleration.

We require that $K$ is compact to perform reachability analysis (Sec. \ref{sec:reachability}).
Let $K_i$ denote the parameters for joint $i$.
For each joint $i$, we specify $K_i = \Kvi \times \Kai$, where
\begin{align}
    \Kvi = \left[\avgkvi - \delkvi,~\avgkvi+\delkvi\right],\quad\Kai = \left[\avgkai - \delkai,~\avgkai+\delkai\right],
\end{align}
with $\avgkvi$, $\avgkai$, $\delkvi$, $\delkai \in \R$ and $\delkvi, \delkai \geq 0$.
To implement acceleration limits (i.e., to bound $\Kai$), we ensure
\begin{align}\label{eq:accel_limit_implementation}
    \Kai = \left[\max\left\{-\ddqlim, \ol{\kai} - \Delta \kai \right\}, \min\left\{\ddqlim, \ol{\kai} + \Delta \kai \right\} \right].
\end{align}

Next, we use these parameterized trajectories to build parameterized reachable sets of joint configurations.

%% file: sections/03_reachability.tex
\section{Offline Reachability Analysis}
\label{sec:reachability}

ARMTD uses short parameterized trajectories of joint angles for trajectory planning.
We now describe a Joint Reachable Set (JRS) containing all such parameterized trajectories.
All computations in this section are performed offline.
\subsubsection{Theory}
Since each $q_i$ represents a rotation, we examine trajectories of $\cos(q_i)$ and $\sin(q_i)$, as shown in Fig. \ref{fig:method_overview}.
By Def. \ref{def:traj_param_generic}, $q(\cdot\,;k)$ is at least once differentiable.
We can write a differential equation of the sine and cosine as a function of the joint trajectory, where $k$ is a constant:
\begin{align}\label{eq:sin_and_cos_diffeq}
    \frac{d}{dt}\begin{bmatrix}
        \cos(q_i(t;k)) \\ \sin(q_i(t;k)) \\ k
    \end{bmatrix} =
    \begin{bmatrix}
        -\sin(q_i(t;k))\dot{q}_i(t;k) \\ \cos(q_i(t;k))\dot{q}_i(t;k) \\ 0
    \end{bmatrix}.
\end{align}
We then define the parameterized JRS of the $i$\ts{th} joint:
\begin{align}\begin{split}\label{eq:cos_sin_reach_set}
    \frs_i = \bigg\{&(c,s,k)\in \R^2\times K\,\mid\,\exists\,t \in T\ \regtext{s.t.}\ q_i \regtext{ as in Def. \ref{def:traj_param_generic}, }\\
    & c = \cos(q_i(t;k)),\ s = \sin(q_i(t;k)),\\
    & \regtext{and } \tfrac{d}{dt}\big(\cos(q_i(t;k)),\sin(q_i(t;k)),k\big)\ \regtext{as in \eqref{eq:sin_and_cos_diffeq}}\bigg\}.
\end{split}\end{align}
We account for different initial joint angles, and use the JRSs to overapproximate the forward occupancy $\FO$, in Sec. \ref{sec:online_planning}.

\subsubsection{Implementation}\label{subsec:reachability_implementation}
We represent \eqref{eq:cos_sin_reach_set} using zonotopes, a subclass of polytopes amenable to reachable set computation \cite{girard2005reachability}.
A \defemph{zonotope} is a set in $\R^n$ in which each element is a linear combination of a \defemph{center} $x \in \R^n$ and \defemph{generators} $g^1,\cdots,g^p \in \R^n,\ p \in \N$:
\begin{align}
\label{eq:zono_long}
Z = \left\lbrace  y \in \mathbb{R}^n \ \Big| \ y = x + \sum_{i = 1}^{p} \bt^i g^i,\ -1 \leq \bt^i \leq 1 \right\rbrace.
\end{align}
We denote $Z = (x, g^i, \sym{\bt^i})^p$ as shorthand for a zonotope with center $x$, a set of generators $\{g^i\}_{i=1}^p$, and a set of indeterminate coefficients $\{\sym{\bt^i}\}_{i=1}^p$ corresponding to each generator.
When an indeterminate coefficient $\sym{\bt^i}$ is \defemph{evaluated}, or assigned a particular value, we write $\bt^i$ (i.e., without angle brackets).

To represent the JRS, we first choose a time step $\Delta t \in \R$ such that $\frac{\tfin}{\Delta t} \in \N$ and partition $T$ into $\frac{\tfin}{\Delta t}$ closed intervals each of length $\Delta t$, indexed by $\N_T = \left\{0,1,\cdots,\tfrac{t_f}{\Delta t}-1\right\}$.
We represent $\frs_i$ with one zonotope per time interval, which is returned by $J_i: \N_T \to \P(\R^2\times K)$.
For example, the zonotope $J_i(n)$ corresponds to the time interval $[n\Delta t, (n+1)\Delta t]$.
We abuse notation and let $t$ index the subinterval of $T$ that contains it, so that $\Zit = J_i\left(\lfloor t/\Delta t\rfloor\right)$ where $\lfloor\cdot\rfloor$ rounds down to the nearest integer.
We use similar notation for the center, generators, and indeterminates.

Next, we make an initial condition zonotope $\Zio \subset \R^2\times K$:
\begin{equation}
\label{eq:CORAinitialSet}
    \Zio = \bigg(\cio, \left\{\givo, \giao\right\}, \left\{\sym{\lkvio},\sym{\lkaio}\right\}\bigg),
\end{equation}
with $\cio = [1, 0, \avgkvi, \avgkai ]^\top$, $\givo = [0, 0, \delkvi, 0 ]^\top$, $\giao = [0, 0, 0, \delkai ]^\top$.
The indeterminates $\sym{\lkvio}$ and $\sym{\lkaio}$ correspond to $\givo$ and $\giao$.
$\Zio$ contains $\Kvi$ and $\Kai$ in the $\kvi$ and $\kai$ dimensions.

Finally, we use an open-source toolbox \cite{althoff_cora} with the time partition, differential equation \eqref{eq:sin_and_cos_diffeq} and \eqref{eq:traj_parameterization}, and initial set $\Zio$ to overapproximate \eqref{eq:cos_sin_reach_set}.
Importantly, by \cite[Thm. 3.3 and Prop. 3.7]{althoff2010reachability}, one can prove the following:
\begin{equation}
    \label{eq:zono_overapprox}
    \frs_i \subseteq \bigcup_{t \in T}\Zit.
\end{equation}


\noindent JRSs are illustrated in Fig. \ref{fig:method_overview}. Next, we use the JRSs online to build an RS for the arm and identify unsafe plans in each receding-horizon iteration.

%% file: sections/04_online_planning.tex
\begin{figure}
    \centering
    \includegraphics[width=0.95\columnwidth]{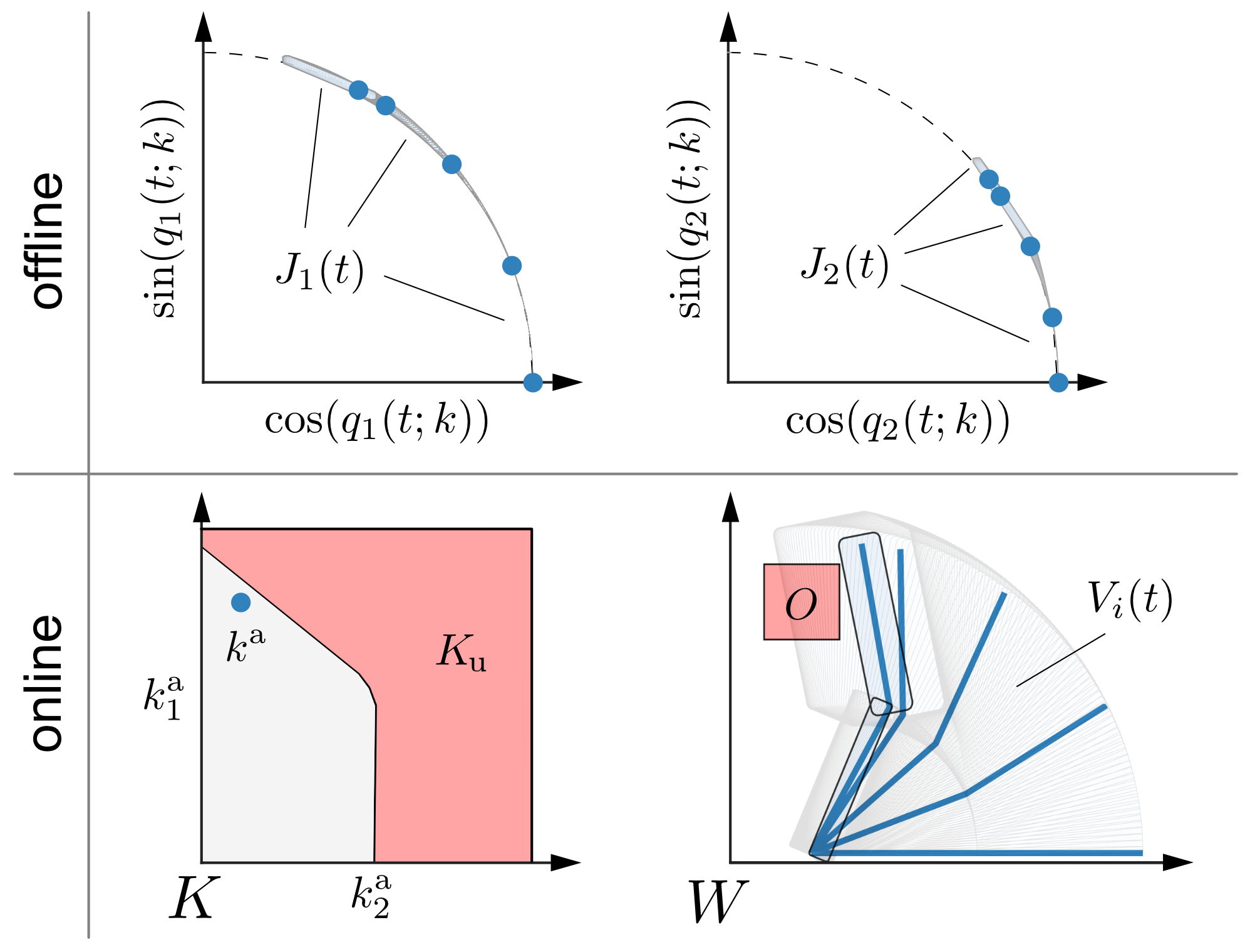}
    \caption{An overview of the proposed method for a 2-D, 2-link arm.
    Offline, ARMTD computes the JRSs, shown as the collection of small grey sets $\Zit$ overlaid on the unit circle (dashed) in the sine and cosine spaces of two joint angles.
    Note that each JRS is conservatively approximated, and parameterized by trajectory parameters $K$.
    Online, the JRSs are composed to form the arm's reachable set $\Vit$ (large light grey sets in $W$), maintaining a parameterization by $K$.
    The obstacle $O$ (light red) is mapped to the unsafe set of trajectory parameters $K\unsafe \subset K$ on the left, by intersection with each $\Vit$.
    The parameter $\ka$ represents a trajectory, shown at five time steps (blue arms in $W$, and blue dots in joint angle space).
    The subset of the arm's reachable set corresponding to $\ka$ is shown for the last time step (light blue boxes with black border), critically not intersecting the obstacle, which is guaranteed because $\ka \not\in K\unsafe$.
    }
    \label{fig:method_overview}
    \vspace*{-0.4cm}
\end{figure}

\section{Online Planning}
\label{sec:online_planning}

We now present ARMTD's online algorithm for a single receding-horizon iteration (see Alg. \ref{alg:online_planning} and Fig. \ref{fig:method_overview}).
First, we construct the parameterized RS of the entire arm from the JRS of each joint.
Second, we identify unsafe trajectory plans.
Third, we optimize over the safe plans to minimize an arbitrary cost function.
If no solution is found, we execute the previous plan's fail-safe maneuver.
Note, we present self-intersection constraints in Appendix \ref{apdx:self_intersection}.

\subsection{Reachable Set Construction}
\subsubsection{Theory}
Recall that ARMTD plans while the robot is executing its previous plan.
Therefore, ARMTD must estimate its future initial condition $(\initq,\initdq) \in Q\times\dot{Q}$ as a result of its previous plan by integrating \eqref{eq:sin_and_cos_diffeq} for $t\plan$ seconds.
At the beginning of each online planning iteration, we use $(\initq,\initdq)$ to compose the RS of the arm from the low-dimensional JRSs.
Denote each link's RS $\linkfrs_i$, formed from all $\frs_j$ with $j \leq i$:
\begin{align}\begin{split}\label{eq:RS_of_link_i}
    \linkfrs_i = \bigg\{&\big(Y, k\big) \in \P(W) \times K \ \Big|\ \exists\,t\in T\ \regtext{s.t.}\\
    &\dot{q}_i(0;k) = \initdqi,\ Y = \FO_i(q(t; k) + \initq),\\
    &\regtext{and}\ \left(\cos(q_j(t; k)), \sin(q_j(t; k)), k\right) \in \frs_j\ \forall\ j \leq i \bigg\} 
\end{split}\end{align}
with $\FO_i$ as in \eqref{eq:forward_occupancy_i}.
Each $\linkfrs_i$ is formed by trajectories which start at the given initial conditions $(\initq,\initdq)$.
The RS of the entire arm, $\linkfrs \subset W\times K$, is then $\linkfrs = \bigcup_i \linkfrs_i$.

\subsubsection{Implementation} \label{sec:online_RS_implementation}

It is important that we overapproximate $\linkfrs$ to guarantee safety when planning.
To do this, we overapproximate $\FO$ for all configurations in each $\frs_i$ (see Alg. \ref{alg:compose_reachable_sets}).

First, we fix $\initdq$ by obtaining subsets of the JRSs containing trajectories with the given initial velocity.
To do so, we note a property of the zonotope JRS:
\begin{lem}\label{lem:one_k_sliceable_gen_per_tope}
There exist $J_i: \N_T \to \P(\R^2\times K)$ that overapproximate $\frs_i$ as in \eqref{eq:zono_overapprox} such that, for each $t \in T$, $\Zit$ has only one generator with a nonzero element, equal to $\delkvi$, in the dimension corresponding to $\kvi$; we denote this generator $\givt$.
Similarly, $\Zit$ has only one generator $\giat$ (distinct from $\givt$) with a nonzero element, $\delkai$, for $\kai$.
\end{lem}
\noindent 
Note, the zonotopes created by the open-source toolbox \cite{althoff_cora} satisfy Lem. \ref{lem:one_k_sliceable_gen_per_tope}.
For each $\Zit$, we denote the center $\cit$, the generators $\{\givt,\giat,\gjt\}$, and the corresponding indeterminates $\left\{\sym{\lkvit},\sym{\lkait},\sym{\btjt}\right\}$ for $j = 1,\cdots, p(t) \in \N$.
We write $p(t)$ since the number of generators is not necessarily the same for each $\Zit$ \cite{althoff_cora}.
For all $t$ except $0$, $\givt$ and $\giat$ may have nonzero elements in the cosine and sine dimensions, due to nonzero dynamics and linearization error.
The generators $\givt$ and $\giat$ are important because they let us obtain a subset of the JRS corresponding to a particular choice of parameters $\kvi$ and $\kai$.
We refer to this operation as \defemph{slicing}, and we call $\givt$ and $\giat$ \defemph{$\kv$-sliceable} and \defemph{$\ka$-sliceable}, respectively.

To this end, we define $\slice$ in Alg. \ref{alg:slice}.
We slice a zonotope by taking in a set of indeterminate coefficients and corresponding values with which to evaluate them.
We evaluate an indeterminate by multiplying its associated generator by the given value.
We then \defemph{remove} the corresponding indeterminate from the set.
Since any zonotope generator has only one indeterminate, once its indeterminate is evaluated, it is called \defemph{fully-sliced}, and added to the center of the zonotope.
Later in this section (Def. \ref{def:rotatotope}), we construct zonotope-like objects called \defemph{rotatotopes}, which have multiple indeterminates per generator (so, a generator could be sliced without being fully-sliced).
For additional explanation of slicing, see Appendix \ref{apdx:explanation:slicing}.

\begin{algorithm}[t]
\small
\begin{algorithmic}[1]
    \State // Let $Z = (x,g^i,\sym{\bt^i})^p$ denote the input zonotope or rotatotope

    \State  $Z_{\regtext{sliced}} \leftarrow (x,g^i,\sym{\bt^i})^p$ // allocate output

    \State{\bf for} $i = 1,\cdots,p$ // iterate over generator/indeterminate pairs
    
    \State\hspace{0.2in}{\bf for} $j = 1,\cdots,n$  // iterate over input values
    
    \State\hspace{0.4in}{\bf if} $\sym{\sg^j} \in \sym{\bt^i}$
    
    \State\hspace{0.6in} $g^i \leftarrow \sg^j g^i$ // multiply generator by value
    
    \State\hspace{0.6in} $\sym{\bt^i} \leftarrow \sym{\bt^i} \setminus \sym{\sg^j}$ // remove evaluated indeterminate

    \State\hspace{0.4in}{\bf end if}
    
    \State\hspace{0.2in}{\bf end for}
    
    \State\hspace{0.2in}{\bf if} $\sym{\bt^i} = \emptyset$ // if fully-sliced, then $g^i$ is no longer needed
    
    \State\hspace{0.4in} $x \leftarrow x + g^i$ and $g^i \leftarrow \emptyset$ // shift center, remove generator
    
    
    \State\hspace{0.2in}{\bf end if}
    
    \State{\bf end for}
    
    \end{algorithmic}
\caption{\small $Z_{\regtext{sliced}} = \slice\left(Z, \{\sym{\sg^j}\}_{j=1}^n, \{\sg^j\}_{j=1}^n\right)$}
\label{alg:slice}
\end{algorithm}

For each joint $i$, recall that each $\Zit$ has generator $\givt$, with indeterminate $\sym{\lkvit}$ and nonzero element $\delkvi$ corresponding to the $\kvi$ dimension.
Also, $\cit$ (the center of $\Zit$) has the value $\avgkvi$ in that same dimension.
We use $\initdq$ to slice each $\Zit$:
\begin{equation}
    \Zitv = \slice\left(\Zit,\ \sym{\lkvit},\ (\initdq - \avgkvi)/\delkvi\right) \label{eq:slice_init_qdot}
\end{equation}
Note, we ensure $\initdq \in K_v$ later in this section.
We denote $\Zitv = (\citv, \big\{\giat,\gjt\big\}, \big\{\sym{\lkait},\sym{\btjt}\big\})^{p(t)}$, 
where $\citv$ is the new (shifted) center and $p(t) \in \N$ is the new number of generators, other than $\giat$, left after slicing.
$\Zitv$ contains a set of $\cos(q_i(t ;k))$ and $\sin(q_i(t ;k))$ reachable for a single value of $\kvi$, but for a range of $\kai$.
Denote the components of $\Zitv$ as $\citv =  [c_i^{\regtext{v}},s_i^{\regtext{v}}, \initdq_i, \avgkai]^\top$, $\giat = [c_i^{\regtext{a}},s_i^{\regtext{a}},0,\Delta{\kai}]^\top$ and $\gjt = [c_i^j,s_i^j,0,0]^\top$ for each $j = 1,...,p(t)$.
Note from Lem. \ref{lem:one_k_sliceable_gen_per_tope} that $c_i^{\regtext{a}}$ and $s_i^{\regtext{a}}$ are generally non-zero, and $\delkai$ is constant.

The forward occupancy map $\FO$ uses rotation matrices formed from the cosine and sine of each joint.
By overapproximating these matrices, we can overapproximate $\FO$.
To this end, we represent sets of rotation matrices with matrix zonotopes.
A \defemph{matrix zonotope} $M \subset \R^{n \times n}$ is a set of matrices parameterized by a center $X$ and generators $G\upone,\cdots, G\upm$:
\begin{equation}
    M = \left\lbrace A \in \R^{n \times n} \ \Big| \ A = X + \sum_{j=1}^m G^j\lambda^j, -1 \leq \lambda^j \leq 1 \right\rbrace.
\end{equation}
We use $M = (X, G^j, \sym{\lm^j})^m$ as shorthand for a matrix zonotope with center $X$, generators $\{G^j\}_{j=1}^m$, and indeterminate coefficients $\{\sym{\lm^j}\}_{j=1}^m$.
Note, superscripts are indices, not exponentiation, of matrix zonotope generators.

We use each sliced zonotope $\Zitv$ to produce a matrix zonotope $\Mit$ that overapproximates the rotation matrices for each joint $i$ at each time $t$.
We do so by reshaping the center and generators of $\Zitv$ (and keeping its indeterminates), then rotating the resulting matrix zonotope by the initial joint angle $\initq$; we call this the \texttt{makeMatZono} function in Alg. \ref{alg:compose_reachable_sets}.
See Appendix \ref{apdx:explanation:mat_zono} for an example of $\Mit$.

Importantly, $\Mit$ satisfies the following property:
\begin{lem} \label{lem:mat_zono_overapprox_R}
For any parameterized trajectory $q: T \to Q$ with  $\kvi = \initdq$, every $R_i(q_i(t;k)) \in M_i(t)$.
\end{lem}

Now we use $\Mit$ to overapproximate the link RS $\linkfrs_i$.
Given the joint displacements $l_i$ and link volumes $L_i$, we specify $l_j \in \R^3$ as a zonotope with center $l_j$ and no generators, and $L_i$ as a zonotope overapproximating the volume of link $i$.
We multiply the matrix zonotopes $\Mit$ by $L_i$ to overapproximate a swept volume, hence the following definition:
\begin{defn}\label{def:rotatotope}
Let $Z = (x, g^i, \sym{\bt^i})^p$ be a zonotope and $M = (X, G^j, \sym{\lm^j})^m$ be a matrix zonotope.
Let $MZ := \{y \in \R^n\ |\ y = Az,\ A \in M,\ z \in Z\} \subset \R^n$.
We call $MZ$ a \defemph{rotatotope}, which can be written:
\begin{align}\begin{split}\label{eq:matrix_zono_times_zono}
    MZ = \bigg\{y \in \R^n\ \mid\ &y = Xx + \tsum_i\,\bt\upi Xg\upi + \tsum_j \lambda\upj G\upj x + \\
    &+\tsum_{i,j}\, \bt\upi\lambda\upj G\upj g\upi ,\ -1 \leq (\bt,\lambda) \leq 1 \bigg\},
\end{split}\end{align}
where $i = 1,\cdots, p$ and $j = 1,\cdots, m$.  
\end{defn}

\noindent We use the shorthand $MZ = \left(\hat{x}, \hat{g}^r, \sym{\gm^r}\right)^s$ where $\hat{x} = Xx$, $s = (p+1)(m+1)-1$, and the generator and coefficient sets are
\begin{align*}
    \{\hat{g}^r\}_{r=1}^{s} &= \{Xg^1, \cdots, Xg^p, G^1x, \cdots, G^mx, G^1g^1, \cdots, G^mg^p\} \\
    \{\sym{\gm^r}\}_{r=1}^{s} &= \{\sym{\bt^1},\cdots,\sym{\bt^p},\sym{\lm^1},\cdots,\sym{\lm^m},\sym{\bt^1\lm^1},\cdots,\sym{\bt^p\lm^m}\}.
\end{align*}
Rotatotopes are a special class of polynomial zonotopes \cite{althoff_cora}.
Each $\sym{\gm^r}$ for $r > p+m$ is a product of indeterminate coefficients from $M$ and $Z$.
For a pair of indeterminate coefficients $\sym{\gm^1}$ and $\sym{\gm^2}$, the notation $\sym{\gm^1\gm^2}$ indicates the product $\sym{\gm^1}\sym{\gm^2}$.
We call $\sym{\gm^1}$ and $\sym{\gm^2}$ the \defemph{factors} of $\sym{\gm^1\gm^2}$.

As noted earlier, we use $\slice$ with rotatotopes, for which we now define removing factors generically.
We denote the \defemph{removal} of the $i$\ts{th} indeterminate coefficient of $\sym{\gm^1\gm^2\cdots\gm^n}$ as:
\begin{align}\label{eq:remove}
    \sym{\gm^1\gm^2\cdots\gm^n}\setminus\sym{\gm^i} = \sym{\gm^1\gm^2\cdots\gm^{i-1}\gm^{i+1}\cdots\gm^n}.
\end{align}
We define $\sym{\gm^1\gm^2\cdots\gm^n}\setminus\sym{\gm^1\gm^2\cdots\gm^n} = \emptyset$.
We write $\sym{\sg} \in \sym{\gm^1\gm^2\cdots\gm^n}$ to denote that $\sym{\sg}$ is a factor of $\sym{\gm^1\gm^2\cdots\gm^n}$. 

Two useful properties follow from the rotatotope definition:

\begin{lem}\label{lem:matzono_times_zono_equals_rotato}
A matrix zonotope times a rotatotope is a rotatotope.
\end{lem}

\begin{lem}\label{lem:mink_sums}
(Zono/rotatotope Minkowski sum)
Consider two zonotopes $X = (x,g_X^i,\sym{\zeta^i})^n$ and $Y = (y,g_Y^j,\sym{\psi^j})^m$.
Then $X\oplus Y = (x+y,\{g_X^i,g_Y^j\},\{\sym{\zeta^i},\sym{\psi^j}\})_{i=1,j=1}^{i=n,j=m}$, which is a zonotope centered at $x+y$ with all the generators and indeterminates of both $X$ and $Y$.
Similarly, for two rotatotopes, $V = (v, g_V^i, \sym{\mu^i})^n$ and $W = (w, g_W^j, \sym{\om^j})^m)$,
\begin{align}\label{eq:mink_sum_rotatotope}
    V \oplus W = \left(v+w, \{g_V^i, g_W^j\}, \{\sym{\mu^i},\sym{\om^j}\}\right)_{i=1,j=1}^{i=n,j=m}.
\end{align}
\end{lem}
\noindent That is, the Minkowski sum is given by the sum of the centers and the union of the generators/indeterminate sets.

We use rotatotopes to overapproximate the forward occupancy map of each link by \defemph{stacking} rotatotopes representing link volume on top of rotatopes representing joint positions:

\begin{lem}
\label{lem:rotatotopes_overapprox_FO}
For any $t \in T$ and $k \in K$, $\FO_i(q(t;k)) \subseteq \Vit$, where
\begin{align}\label{eq:stacking}
    \Vit = \bigoplus_{j < i} \Bigg(\prod_{n \leq j}\Mnt\,\{l_j\}\Bigg) \oplus \left(\prod_{n\leq i}\Mnt L_i\right) \subset W.
\end{align}
\end{lem}
\noindent Lem. \ref{lem:rotatotopes_overapprox_FO} lets us overapproximate the RS: $\linkfrs_i \subseteq \bigcup_{t \in T}\Vit \implies \linkfrs \subseteq \bigcup_{t, i} \Vit$, as shown in Fig. \ref{fig:method_overview}.
Alg. \ref{alg:compose_reachable_sets} computes $\Vit$ (see Appendix \ref{apdx:explanation:reduction} for further computational details).

Though $\Vit \subset W$, many of its generators are $\ka$-sliceable, because they are the product of $\ka$-sliceable matrix zonotope generators.
Denote $\Vit = (\cithat, \hat{g}_i^j(t), \sym{\hat{\bt}_i^j(t)})^{p(t)}$.
Formally, the $j$\ts{th} generator $\hat{g}_i^j(t)$ is $\ka$-sliceable if there exists at least one $\sym{\lkant} \in \sym{\hat{\bt}_i^j(t)}$ with $n \leq i$.
This means, by slicing by $\ka$, we can obtain a subset of $\Vit$ corresponding to that parameter.
We make the distinction that a generator $\hat{g}_i^j(t)$ is \textit{fully-$\ka$-sliceable} if \emph{all} of its indeterminates are evaluated when sliced by $\ka$, i.e. $\sym{\hat{\bt}_i^j(t)} \subseteq \bigcup_{n \leq i}\sym{\lkant}$.
Fully-$\ka$-sliceable generators are created by multiplying $\ka$-sliceable generators with each other or with centers in \eqref{eq:stacking}. 
These generators are important because all of their indeterminates are evaluated by the trajectory optimization decision variable $\ka$, which we use in Sec. \ref{sec:online_constraint_generation}.


\begin{algorithm}[t]
\small
\begin{algorithmic}[1]
\State{\bf parfor} $t \in T$ // parallel for each time step

\State\hspace{0.2in}{\bf for} $i = 1:n_q$ // for each joint

    \State\hspace{0.4in}$\lkvit \leftarrow (\initdq - \avgkvi)/(\delkvi)$ // get value for \eqref{eq:slice_init_qdot}

    \State\hspace{0.4in}$\Zitv \leftarrow \texttt{slice}(\Zit, \sym{\lkvit}, \lkvit)$ // slice JRS

    \State\hspace{0.4in}$\Mit \leftarrow \texttt{makeMatZono}(\Zitv,\initq)$ 
    
    \State\hspace{0.4in}$\Vit \leftarrow \Mit L_i$ // init $\Vit$ for link volume RS
    
    \State\hspace{0.4in}$\Uit \leftarrow l_{i-1}$ // init rotatotope for joint location
    
    \State\hspace{0.4in}{\bf for} $j = (i-1):-1:1$ // predecessor joints
    
        \State\hspace{0.6in}$\Vit \leftarrow M_j^t \Vit$ // rotate link volume
        
        \State\hspace{0.6in}$\Uit \leftarrow M_j^t \Uit$ // rotate joint location
    
        
    \State\hspace{0.4in}{\bf end for}

    \State\hspace{0.4in}{\bf for} $j = (i-1):-1:1$ // predecessor joints

        \State\hspace{0.6in}$\Vit \leftarrow \Vit \oplus \Ujt$ // stack link on joints
        
    \State\hspace{0.4in}{\bf end for}
    
    \State\hspace{0.2in}{\bf end for}
    
\State{\bf end parfor}



    
    

\end{algorithmic}
\caption{\small $\{\Vit\,:\, i = 1,\cdots,n_q,\ t\in T\} = \texttt{composeRS}(\initq,\initdq)$}
\label{alg:compose_reachable_sets}
\end{algorithm}

\subsection{Constraint Generation}
\subsubsection{Theory}
With the RS composed, we now use $\linkfrs$ to find all unsafe trajectory parameters $k \in K\unsafe \subseteq K$ that could cause collisions with obstacles.
We treat $K\unsafe$ as a constraint for trajectory optimization, shown in Fig. \ref{fig:method_overview}.
Recall $\qlim^-$, $\qlim^+$, and $\dqlim$ are joint limits.
Let $\obsset$ be a set of obstacles.
At each planning iteration, the unsafe trajectory parameters are $K\unsafe = K\jlim \cup K\obs$, where
\begin{align}
\begin{split}K\jlim &= \big\{k\ |\ \exists\ t \in T\ \regtext{s.t.} \ q(t;k) < \qlim^- \ \regtext{or}\ q(t;k) > \qlim^+ \label{eq:K_lim}\\
&\quad\quad\quad\regtext{or}\ |\dot{q}(t;k)| > \dqlim \big\}\end{split}\\
    K\obs &= \big\{k\ |\ Y \cap O\neq \emptyset,\, (Y,k) \in \linkfrs,\, O \in \obsset\big\}.\label{eq:K_obs}
\end{align}

\subsubsection{Implementation}\label{sec:online_constraint_generation}
We represent $K\jlim$ with functions $h\jlim: \Ka \to \R$.
Notice in \eqref{eq:traj_parameterization} that $q(t; k)$ is piecewise quadratic in $k$ and $\dot{q}(t;k)$ is piecewise linear in $k$, so the parameterized trajectory extrema can be computed analytically.
We construct $h\jlim$ from $\initdqi$, $\qlim$, and $\dqlim$, such that $h\jlim(\ka) < 0$ when feasible.

To represent $K\obs$ (depicted in Fig. \ref{fig:method_overview}), first consider a particular $\ka$.
We test if the corresponding subset of each rotatotope $\Vit$ could intersect any obstacle $O \in \obsset$.
We overapproximate each $O$ by a zonotope, which is always possible for compact, bounded sets \cite{althoff2010reachability} that appear in common obstacle representations such as octrees \cite{meagher1982geometric} or convex polytopes \cite{lien2007approximate}.
To proceed, we must test if two zonotopes intersect:
\begin{lem}\label{lem:zono_int}
\cite[Lem. 5.1]{guibas2003zonotopes}
For two zonotopes $X = (x, g^i, \sym{\bt^i})^n$ and $Y = (y, g^j, \sym{\bt^j})^m$, $X \cap Y \neq \emptyset$ iff $y$ is in the zonotope $X\buf = (x, g^i, \sym{\bt^i})^n \oplus (0, g^j, \sym{\bt^j})^m$, where the subscript indicates $X$ is buffered by the generators of $Y$.
\end{lem}

\noindent Since zonotopes are convex polytopes \cite{guibas2003zonotopes}, by \cite[Theorem 2.1]{althoff2010reachability}, one can implement Lem. \ref{lem:zono_int} by computing a \defemph{half-space representation} $(A\buf,b\buf)$ of $X\buf$ for which $A\buf z - b\buf \leq 0 \iff z \in X\buf$, where the inequality is taken elementwise.
Using this representation, $X \bigcap Y = \emptyset \iff \max(A\buf y - b\buf) > 0$.
We can use Lem. \ref{lem:zono_int} for collision avoidance by replacing $X$ (resp. $Y$) with a zonotope representing the arm (resp. an obstacle).


However, since we use rotatotopes, we need the following:
\begin{lem}\label{lem:zono_overapprox_rotatotope}
Any rotatotope $MZ$ as in \eqref{eq:matrix_zono_times_zono} can be overapproximated by a zonotope.
\end{lem}
\noindent 
So, we can overapproximate the intersection of each $\Vit$, sliced by $\ka$, with each $O \in \obsset$.
Note, we only slice the fully-$\ka$-sliceable generators of $\Vit$, and treat all other generators conservatively by applying Lemma \ref{lem:zono_overapprox_rotatotope}.
That is, we do not slice any generators that have any indeterminates in addition to $\sym{\lkait}$, and instead use those generators to (conservatively) buffer obstacles.

To check intersection, we separate $\Vit$ into two rotatotopes,
\begin{align}\label{eq:vit_split_slice_and_buff}
    \Vitslc = \left(\cit, g\slc^j, \sym{\kp\slc^j}\right)\ \regtext{and}\ \Vitbuf = \left(0, g\buf^n, \sym{\bt\buf^n}\right),
\end{align}
such that $\Vit = \Vitslc \oplus \Vitbuf$, where $\Vitslc$ has only fully-$\ka$-sliceable generators.
That is, each $\sym{\kp\slc^j}$ is a product of \textit{only} $\sym{\lkait}$ for one or more $i \in \{1,\cdots,n_q\}$.
Note, the number of generators/indeterminates in $\Vitslc$ and $\Vitbuf$ is omitted to ease notation.
For any $\ka \in \Ka$, since every generator of $\Vitslc$ is $\ka$-sliceable, slicing $\Vitslc$ by $\ka$ returns a point.
We express this with $\eval: \P(W) \times \Ka \to \R^3$ for which
\begin{align}\label{eq:eval}
    \eval(\Vitslc, \ka) = \slice\left(\Vitslc,\big\{\sym{\lkait}\big\}_{i=1}^{n_q},\{\kp(i)\}_{i=1}^{n_q}\right)
\end{align}
where $\kp(i) = (\kai - \avgkai)/\delkai$.
Note, $\eval$ can be implemented as the evaluation of polynomials.

Now, let $A\obs$ and $b\obs$ be the halfspace representation of $O\buf = O \oplus \Vitbuf$, and let $x = \eval(\Vitslc,\ka)$.
Then,
\begin{equation} \label{eq:sliced_rotato_int_obs}
    \left(\{x\} \oplus \Vitbuf \right) \cap O = \emptyset \iff -\max\{ A\obs x - b\obs\} < 0
\end{equation}
where $\{x\} \oplus \Vitbuf$ is overapproximated as a zonotope by applying Lem. \ref{lem:zono_overapprox_rotatotope}.
We use \eqref{eq:sliced_rotato_int_obs} to overapproximate the parameters $K\obs$ \eqref{eq:K_obs} with $h\obs: \N \times T\times \obsset\times\Ka \to \R$ for which
\begin{align}
    \label{eq:h_obs}
    h\obs(*,\ka) = -\max\big\{A\obs(*)\eval(\Vitslc,\ka) - b\obs(*)\big\}.
\end{align}
where $* = (i,t,O)$ for space.
Here, $A\obs(i, t, O)$ and $b\obs(i, t, O)$ return the halfspace representation of $O \oplus \Vitbuf$.
Importantly, for each obstacle, time, and joint, $h\obs$ is a max of a linear combination of polynomials in $\ka$ (per \eqref{eq:eval} and Alg. \ref{alg:slice}), so we can take its subgradient with respect to $\ka$ \cite{boyd2003subgradient} (also see \cite[Thm. 5.4.5]{polak2012optimization}).
This constraint conservatively approximates $K\obs$:
\begin{lem}\label{lem:hobs_is_conservative}
If $\ka \in K\obs$, then there exists $i \in \N$, $t \in T$, and $O \in \obsset$ such that $h\obs(i,t,O,\ka) \geq 0$.
\end{lem}

\subsection{Trajectory Optimization}
\subsubsection{Theory}
ARMTD performs trajectory optimization over $K \setminus K\unsafe$ for an arbitrary user-specified cost function $\costfunc: K \to \R$ (which encodes information such as completing a task).
ARMTD attempts to solve the following within $t\plan$:
\begin{align}\label{prog:trajopt_general}
    k\opt = \regtext{argmin}_{k}\big\{\costfunc(k)\ |\ k \not\in K\unsafe\big\}.
\end{align}
If no solution is found in time, the robot tracks the fail-safe maneuever from its previous plan.

\subsubsection{Implementation}\label{sec:online_trajopt}
We implement \eqref{prog:trajopt_general} as a nonlinear program, denoted \texttt{optTraj} in Alg. \ref{alg:online_planning}.
\begin{align}\label{prog:trajopt}
    \underset{\ka\,\in\,\Ka}{\regtext{argmin}}\left\{\costfunc(\ka)\ |\ h\obs(i,t,O,\ka) < 0,\ h\jlim(\ka) < 0\right\}
\end{align}
where the constraints hold for all $i \in \{1,\cdots,n_q\},\ t \in T,\ O \in \obsset$.

\begin{thm}\label{thm:constraints_are_conservative}
Any feasible solution to \eqref{prog:trajopt} parameterizes a trajectory that is collision-free and obeys joint limits over the time horizon $T$.
\end{thm}

ARMTD uses Alg. \ref{alg:online_planning} at each planning iteration.
If the arm does not start in collision, this algorithm ensures that the arm is always safe (see Appendix \ref{apdx:safe_planning}, also see \cite[Remark 70]{kousik2018_RTD_ijrr} or \cite[Theorem 1]{Hauser2012_receding_horizon}).



\begin{algorithm}[t]
\small
\begin{algorithmic}[1]
    
    \State $\{\Vit\} \leftarrow \texttt{composeRS}(\initq,\initdq)$ // Sec. \ref{sec:online_RS_implementation} \label{lin:construct_frs}
    
    \State $(h\obs,h\jlim) \leftarrow \texttt{makeCons}(\initq,\initdq,\obsset,\{\Vit\})$ // Sec. \ref{sec:online_constraint_generation}
    
    \State // solve \eqref{prog:trajopt} within $t\plan$ or else return $q\prev$
    
    \State $q\plan \leftarrow \texttt{optTraj}\left(\costfunc, h\obs, h\jlim, t\plan, q\prev\right)$ // Sec. \ref{sec:online_trajopt} \label{lin:trajopt}

    \end{algorithmic}
\caption{\small $q\plan = \texttt{makePlan}(\initq,\initdq,q\prev,\obsset,\costfunc)$}
\label{alg:online_planning}
\end{algorithm}

%% file: sections/05_demos.tex
\section{Demonstrations}
\label{sec:demos}

We now demonstrate ARMTD in simulation and on hardware using the Fetch mobile manipulator (Fig. \ref{fig:fetch_intro}).
ARMTD is implemented in MATLAB, CUDA, and C++, on a 3.6 GHz computer with an Nvidia Quadro RTX 8000 GPU. 
See our video: \href{https://youtu.be/ySnux2owlAA}{\texttt{youtu.be/ySnux2owlAA}}.
Our code is available: \href{https://github.com/ramvasudevan/arm_planning}{\texttt{github.com/ramvasudevan/arm\_planning}}.

\subsection{Implementation Details}

\subsubsection{Manipulator}
The Fetch arm has $7$ revolute DOFs \cite{wise2016fetch}.
We consider the first $6$ DOFs, and treat the body as an obstacle.
The $7$\ts{th} DOF controls end effector orientation, which does not affect the volume used for collision checking.
We command the hardware via ROS \cite{quigley2009ros} over WiFi.

\subsubsection{Comparison}
To assess the difficulty of our simulated environments, we ran CHOMP \cite{chomp} via MoveIt \cite{moveit} (default settings, straight-line initialization).
We emphasize that CHOMP is not a receding-horizon planner \cite{moveit}; it attempts to find a plan from start to goal with a single optimization program.
However, CHOMP provides a useful baseline to measure the performance of ARMTD.
To the best of our knowledge, no open-source, real-time receding-horizon planner is available for a direct comparison.
Note, we report solve times to illustrate that ARMTD is real-time feasible, but the goal of ARMTD is not to solve as fast as possible; instead, we care about finding provably collision-free trajectories in the allotted time $t\plan$.


\subsubsection{High-level Planner}
Recall that ARMTD performs trajectory optimization using an arbitrary user-specified cost function.
In this work, in each planning iteration, we create a cost function for ARMTD using an intermediate waypoint between the arm's current configuration and a global goal.
These waypoints are generated by a high-level planner (HLP).
Note, the RS and safety constraints generated by ARMTD are independent of the HLP, which is only used for the cost function.
To illustrate that ARMTD can enforce safety, we use two different HLPs, neither of which is guaranteed to generate collision-free waypoints.
First, a straight-line HLP that generates waypoints along a straight line between the arm and a global goal in configuration space.
Second, an RRT* \cite{karaman2011sampling} that only ensures the arm's end effector is collision-free.
Thus, \textbf{ARMTD can act as a safety layer on top of RRT*}.
Note, we allot a portion of $t\plan$ to the HLP in each iteration, and give ARTMD the rest of $t\plan$.
We cannot use CHOMP as a receding-horizon planner with these HLP waypoints, because it requires a collision-free goal configuration. 
For further discussion of the comparison to CHOMP, see Appendix \ref{apdx:explanation:seeding_chomp}.

\subsubsection{Algorithm Implementation}
Alg. \ref{alg:compose_reachable_sets} runs at the start of each ARMTD planning iteration.
We use a GPU with CUDA to execute Alg. \ref{alg:compose_reachable_sets} in parallel, taking approximately $10$--$20$ ms to compose a full RS.
The constraint generation step in Alg. \ref{alg:online_planning} is also parallelized across obstacles and time steps (this takes approximately $10$--$20$ ms for $20$ obstacles).

We solve ARMTD's trajectory optimization \eqref{prog:trajopt} using IPOPT \cite{wachter2006implementation}.
The cost function $\costfunc$ is $||q(\tfin ;k) - q\des||_{2}^{2}$, where $q\des$ is the waypoint specified by the HLP (straight-line or RRT*) at each planning iteration.
We compute analytic gradients/sub-gradients of the cost function and constraints, and evaluate the constraints in parallel.
IPOPT takes $100$--$200$ ms when it finds a feasible solution in a scene with $20$ random obstacles.

\subsubsection{Hyperparameters}
To reduce conservatism, we partition $\Kvi$ into $n_{\regtext{JRS}} \in \N$ equally-sized intervals and compute one JRS for each interval.
At runtime, for each joint, we pick the JRS containing the initial speed $\initdq_i$.
In each JRS, we set $\Delta \kai = \max\left\{ r_{a_2},\ r_{a_1}|\overline{\kvi}|\right\}$, with $r_{a_1}, r_{a_2} > 0$ so the range of accelerations scales with the absolute value of the mean velocity of each JRS.
This reduces conservativism at low speeds, improving maneuverability near obstacles.

We also use these values: $t\plan = 0.5$ s, $\tfin = 1.0$ s, $\Delta t = 0.01$ s, $n_{\regtext{JRS}} = 400$, $\dqlim = \pi \frac{\regtext{rad}}{s}$, $\ddqlim = \pi/3 \frac{\regtext{rad}}{s^2}$, $\overline{\kai} = 0 \frac{\regtext{rad}}{s^2}$, $r_{a_1} = 1/3 s^{-1}$, and $r_{a_2} = \pi/24 \frac{\regtext{rad}}{s^2}$.
For collision checking, we overapproximate the Fetch's links with cylinders of radius 0.146 m.
For further discussion of design choices and hyperparamters, see Appendix \ref{apdx:explanation:design}.

\subsection{Simulations}

\subsubsection{Setup}

\begin{figure}[t]
    \centering
    \includegraphics[width=\columnwidth]{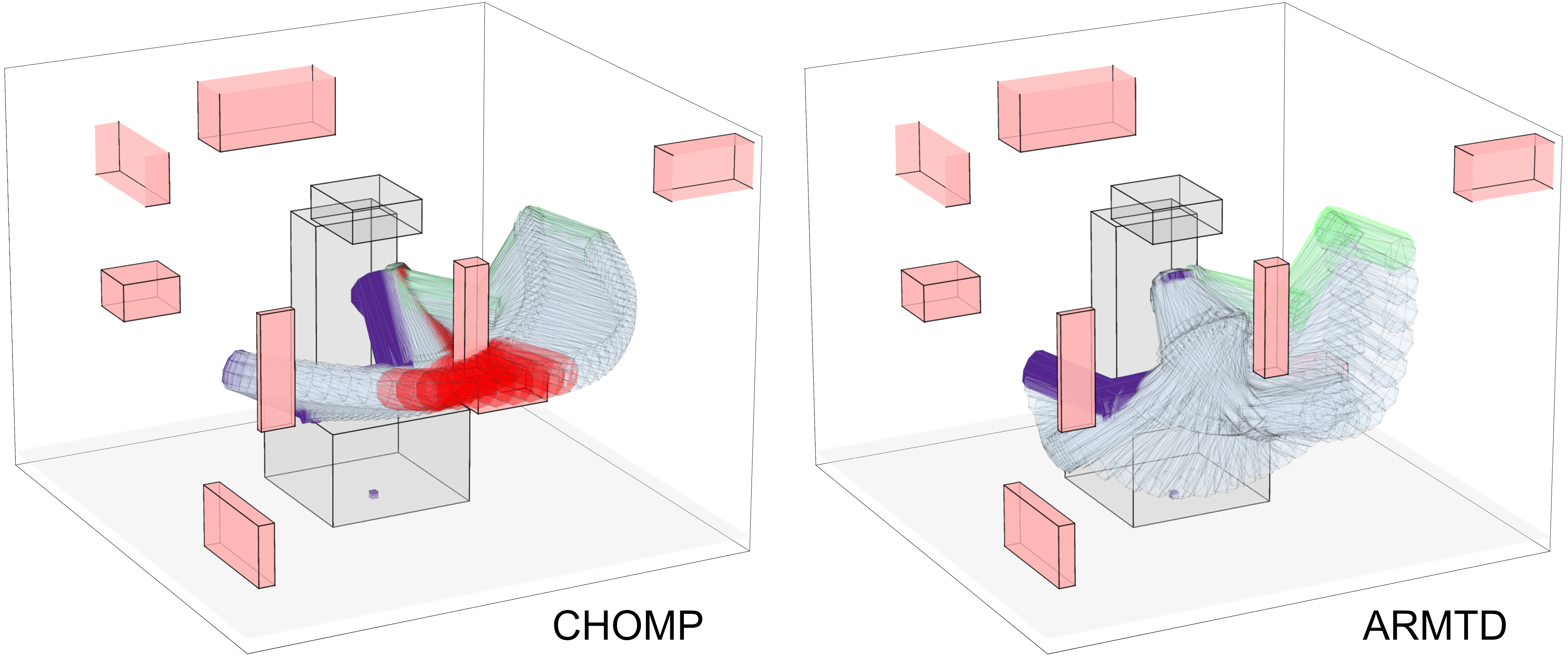}
    \caption{A Random Obstacles scene with $8$ obstacles in which CHOMP \cite{chomp} converged to a trajectory with a collision (collision configurations shown in red), whereas ARMTD successfully navigated to the goal (green); the start pose is shown in purple.
    CHOMP fails to move around a small obstacle close to the front of the Fetch.}
    \label{fig:armtd_chomp_random_scene}
    \vspace*{-0.4cm}
\end{figure}

We created two sets of scenes.
The first set, Random Obstacles, shows that ARMTD can handle arbitrary tasks (see Fig. \ref{fig:armtd_chomp_random_scene}).
This set contains 100 tasks with random (but collision-free) start and goal configurations, and random box-shaped obstacles.
Obstacle side lengths vary from $1$ to $50$ cm, with 10 scenes for each $n_O = 4, 8, ..., 40$.

The second set, Hard Scenarios, shows that ARMTD guarantees safety where CHOMP converges to an unsafe trajectory.
There are seven tasks in the Hard Scenarios set: (1) from below to above a table, (2) from one side of a wall to another, (3) between two vertical posts, (4) from one set of shelves to another, (5) from inside to outside of a box on the ground, (6) from a sink to a cupboard, (7) through a small window.
These scenarios are shown in Fig. \ref{fig:hard_scenarios} in Appendix \ref{apdx:explanation:hard_scenarios}.

\subsubsection{Results}
Table \ref{tab:random_obstacles_results} presents ARMTD (with a straight-line HLP) and CHOMP's results for the Random Obstacles scenarios.
ARMTD reached $84/100$ goals and had $0/100$ crashes, meaning ARMTD stopped safely $16/100$ times without finding a new safe trajectory.
CHOMP reached $82/100$ goals and had $18/100$ crashes.
CHOMP always finds a trajectory, but not necessarily a collision-free one; it can converge to infeasible solutions because it considers a non-convex problem with obstacles as areas of high cost (not as hard constraints).
We did not attempt to tune CHOMP to only find feasible plans (e.g., by buffering the arm), since this incurs a tradeoff between safety and performance.
Note, in MoveIt, infeasible CHOMP plans are not executed (if detected by an external collision-checker).

We report the mean solve time (MST) of ARMTD over all planning iterations, while the MST for CHOMP is the mean over all 100 tasks.
Directly comparing timing is not possible since ARMTD and CHOMP use different planning paradigms; we report MST to confirm ARMTD is capable of real-time planning (note that that ARMTD's MST is less than $t\plan = 0.5$).

We also report the mean normalized path distance (MNPD) of the plans produced by each planner (the mean is taken over all 100 tasks).
The normalized path distance is a path's total distance (in configuration space), divided by the distance between the start and goal.
For example, the straight line from start to goal has a (unitless) normalized path distance of $1$.
ARMTD's MNPD is $24\%$ smaller than CHOMP's, which may be because CHOMP's cost rewards path smoothness, whereas ARMTD's cost rewards reaching an intermediate waypoint at each planning iteration (note, path smoothness could be included in ARMTD's cost function).

Table \ref{tab:hard_scenarios_results} presents results for the Hard Scenarios.
With the straight-line HLP, ARMTD does not complete any of the tasks but also has no collisions.
With the RRT* HLP \cite{karaman2011sampling}, ARMTD completes $5/7$ scenarios.
CHOMP converges to trajectories with collisions in all of the Hard Scenarios.

\subsection{Hardware}
See our video: \href{https://youtu.be/ySnux2owlAA}{\texttt{youtu.be/ySnux2owlAA}}.
ARMTD completes arbitrary tasks while safely navigating the Fetch arm around obstacles in scenarios similar to Hard Scenarios (1) and (4).
We demonstrate real-time planning by suddenly introducing obstacles (a box, a vase, and a quadrotor) in front of the moving arm.
The obstacles are tracked using motion capture, and treated as static in each planning iteration.
Since ARMTD performs receding-horizon planning, it can react to the sudden obstacle appearance and continue planning without crashing.

\begin{table}[t]
\centering
\begin{tabular}{r|c|c|c|c|}
\multicolumn{1}{c|}{\textbf{Random Obstacles}} & \% goals & \% crashes & MST {[}s{]} & MNPD \\ \hline
ARMTD + SL & 84 & 0 & 0.273 & 1.076\\ \hline
CHOMP & 82 & 18 & 0.177 & 1.511\\ \hline
\end{tabular}
\caption{MST is mean solve time (per planning iteration for ARMTD with a straight-line planner, total for CHOMP) and MNPD is mean normalized path distance. MNPD is only computed for trials where the task was successfully completed, i.e. the path was valid.}
\label{tab:random_obstacles_results}
\end{table}

\begin{table}[t]
\centering
\begin{tabular}{r|l|l|l|l|l|l|l|}
\multicolumn{1}{c|}{\textbf{Hard Scenarios}} & \multicolumn{1}{c|}{1} & \multicolumn{1}{c|}{2} & \multicolumn{1}{c|}{3} & \multicolumn{1}{c|}{4} & \multicolumn{1}{c|}{5} & \multicolumn{1}{c|}{6} & \multicolumn{1}{c|}{7} \\ \hline
ARMTD + SL & S & S & S & S & S & S & S\\ \hline
ARMTD + RRT* & O & O & O & S & O & S & O \\ \hline
CHOMP & C & C & C & C & C & C & C \\ \hline
\end{tabular}
\caption{
Results for the seven Hard Scenario simulations.
ARMTD uses straight-line (SL) and RRT* HLPs.
The entries are ``O'' for task completed, ``C'' for a crash, or ``S'' for stopping safely without reaching the goal.
}
\vspace*{-0.25cm}
\label{tab:hard_scenarios_results}
\end{table}

%% file: sections/06_conclusion.tex
\section{Conclusion}\label{sec:conclusion}

This work proposes ARMTD as a real-time, receding-horizon manipulator trajectory planner with safety guarantees.
The method proposes novel reachable sets for arms, which enable safety.
ARMTD can enforce safety on top of an unsafe path planner such as RRT*, shown in both simulation and on hardware.
Of course, ARMTD has limitations: it may not perform in real time without parallelization, is only demonstrated on 6-DOF planning problems, and has not yet been demonstrated planning around humans.
However, because ARMTD uses time-varying reachable sets, it can readily extend to dynamic environments, uncertainty such as tracking error, and planning with grasped objects.
The results in this work show promise for practical, safe robotic arm trajectory planning.


%% file: sections/appendix_supplement.tex
\newtheorem{innercustomthm}{Theorem}
\newenvironment{customthm}[1]
  {\renewcommand\theinnercustomthm{#1}\innercustomthm}
  {\endinnercustomthm}

\newtheorem{innercustomlem}{Lemma}
\newenvironment{customlem}[1]
  {\renewcommand\theinnercustomlem{#1}\innercustomlem}
  {\endinnercustomlem}

\section{Proofs} \label{apdx:proofs}

Here, we provide the proof of each mathematical claim in the paper, plus a short explanation of how each claim is useful.

First, we examine the structure of the JRS zonotope representation.
This structure enables the creation of \emph{fully-$k$-sliceable} generators when we use the JRS to produce rotatotopes.
That is, this lemma enables us to slice the arm's RS to find subsets corresponding to particular trajectory parameters.
\begin{customlem}{\ref{lem:one_k_sliceable_gen_per_tope}}
There exist $J_i: \N_T \to \P(\R^2\times K)$ that overapproximate $\frs_i$ as in \eqref{eq:zono_overapprox} such that, for each $t \in T$, $\Zit$ has only one generator with a nonzero element, equal to $\delkvi$, in the dimension corresponding to $\kvi$, and only one (distinct) generator with a nonzero element, $\delkai$, for $\kai$.
\end{customlem}
\begin{proof}
Given $\Zio$, the subsequent zonotope $\Zi(\Delta t)$ is computed as $\Zi(\Delta t) = e^{F\Delta T}\Zio + E$, where $F$ is found by linearizing the dynamics \eqref{eq:sin_and_cos_diffeq} at $t = 0$ and $E$ is a set that overapproximates the linearization error and the states reached over the interval $[0,\Delta t]$ \cite[Section 3.4.1]{althoff2010reachability}.
This linearized forward-integration and error-bounding procedure is applied to $\Zi(\Delta t)$ to produce $\Zi(2\Delta t)$, and so on, to compute all $\Zit$ in \eqref{eq:zono_overapprox}.
Since $\dot{k} = 0$, we have that $e^{F\Delta}\givo$ equals $\givo$ in the $k$ dimensions (and therefore each $\givt$ does as well, and similarly for $\giat$). 
Since the zero dynamics have no linearization error, one can define $E$ to have zero volume in the $k$ dimensions \cite[Proposition 3.7]{althoff2010reachability}, meaning no generator of any $\Zit$ has a nonzero element in the $k$ dimensions, except for $\givt$ and $\giat$ (which are defined with such nonzero elements).
\end{proof}

We now note that all rotation matrices are contained in the matrix zonotopes $\Mit$, which are produced by slicing and reshaping the JRS zonotopes.
This enables us to conservatively approximate the forward occupancy map.

\begin{customlem}{\ref{lem:mat_zono_overapprox_R}}
For any parameterized trajectory $q: T \to Q$ with  $\kvi = \initdq$, every $R_i(q_i(t;k)) \in M_i(t)$.
\end{customlem}
\begin{proof}
By \cite[Thm. 3.3 and Prop. 3.7]{althoff2010reachability}, all values attained by the sines and cosines of the joint angles are contained in each $\Zit$.
By Alg. \ref{alg:slice} and \eqref{eq:slice_init_qdot}, each $\Zitv$ only contains the values of sine and cosine of $q(t;k)$ for which $\kvi = \initdq$.
Since $\Mit$ only reshapes $\Zitv$, the proof is complete.
\end{proof}

The following lemma confirms that the product of multiple matrix zonotopes times a zonotope is still a rotatotope.
This is necessary to overapproximate the forward occupancy map, wherein the arm's joint rotation matrices are multiplied together (and, analogously, the matrix zonotopes are multiplied together).

\begin{customlem}{\ref{lem:matzono_times_zono_equals_rotato}}
A matrix zonotope times a rotatotope is a rotatotope.
\end{customlem}
\begin{proof}
This follows from the rotatotope definition.
\end{proof}

We use the Minkowski sums of zonotopes and rotatotopes to enable stacking, which is how we build an RS of the entire arm from the low-dimensional JRSs.
We also use the Minkowski sum to dilate obstacles, which is necessary to check for intersection with our arm's RS per Lem. \ref{lem:zono_int} (which we do not prove here, as it is proven in \cite{guibas2003zonotopes}).

\begin{customlem}{\ref{lem:mink_sums}}[Minkowski sum of zonotopes and rotatotopes]
Consider two zonotopes $X = (x,g_X^i,\sym{\zeta^i})^n$ and $Y = (y,g_Y^j,\sym{\psi^j})^m$.
Then $X\oplus Y = (x+y,\{g_X^i,g_Y^j\},\{\sym{\zeta^i},\sym{\psi^j}\})_{i=1,j=1}^{i=n,j=m}$, which is a zonotope centered at $x+y$ with all the generators and indeterminates of both $X$ and $Y$.
Similarly, for two rotatotopes, $V = (v, g_V^i, \sym{\mu^i})^n$ and $W = (w, g_W^j, \sym{\om^j})^m)$,
\begin{align}
    V \oplus W = \left(v+w, \{g_V^i, g_W^j\}, \{\sym{\mu^i},\sym{\om^j}\}\right)_{i=1,j=1}^{i=n,j=m}.\tag{\ref{eq:mink_sum_rotatotope}}
\end{align}
\end{customlem}
\begin{proof}
This follows from the zonotope definition \eqref{eq:zono_long} and rotatotope definition (Def. \ref{def:rotatotope}).
\end{proof}

The following lemma confirms that the sliced and stacked rotatotopes $\Vit$ overapproximate the forward occupancy map $\FO$ for each $i$\ts{th} link.

\begin{customlem}{\ref{lem:rotatotopes_overapprox_FO}}
For any $t \in T$ and $k \in K$, $\FO_i(q(t;k)) \subseteq \Vit$, where
\begin{align}
    \Vit = \bigoplus_{j < i} \Bigg(\prod_{n \leq j}\Mnt\,\{l_j\}\Bigg) \oplus \left(\prod_{n\leq i}\Mnt L_i\right) \subset W.\tag{\ref{eq:stacking}}
\end{align}
\end{customlem}
\begin{proof}
First, note \eqref{eq:stacking} is defined analogously to \eqref{eq:forward_occupancy_i}.
We have $R_i(q_i(t;k)) \in \Mit$ from Lem. \ref{lem:mat_zono_overapprox_R}.
The product of matrix zonotopes multiplied by a zonotope is a rotatope by Def. \ref{def:rotatotope}, and the Minkowski sum of rotatotopes are given exactly using Lem. \ref{lem:mink_sums}.
Therefore, all sets and operations in \eqref{eq:stacking} are exact or conservative (note, we can overapproximate $L_i$ with a zonotope), so $\FO_i(q(t ;k)) \subseteq \Vit$.
\end{proof}

We use this next lemma to overapproximate the swept volume of the arm (represented with rotatotopes); the overapproximation means that ARMTD is provably conservative, which enables safety guarantees.

\begin{customlem}{\ref{lem:zono_overapprox_rotatotope}}
Any rotatotope $MZ$ as in \eqref{eq:matrix_zono_times_zono} can be overapproximated by a zonotope.
\end{customlem}
\noindent \begin{proof}
Consider the components of the indeterminate coefficients of $MZ = (x,g^r,\sym{\gm^r})^s$ that can be written as $\sym{\bt^i\lm^j}$.
When evaluated, $\bt^i\lm^j \in [-1,1]$.
Consider a zonotope $\hat{Z} = (x,g^r,\sym{\sg^r})^s$ with the same center and generators as $MZ$, but where each product $\sym{\bt^i\lm^j}$ is replaced with a single new symbolic coefficient $\sym{\sg^r}$.
If $z \in MZ$, $\exists\ \sg^r \in [-1,1]$ such that $z \in \hat{Z}$.
\end{proof}

The following lemma confirms that our unsafe set representation (the function $h\obs$) is conservative.

\begin{customlem}{\ref{lem:hobs_is_conservative}}
If $\ka \in K\obs$, then there exists $i \in \N$, $t \in T$, and $O \in \obsset$ such that $h\obs(i,t,O,\ka) \geq 0$.
\end{customlem}
\begin{proof}
This follows from Lems. \ref{lem:rotatotopes_overapprox_FO} and \ref{lem:zono_int}; $h\obs$ is positive when the zonotope produced by slicing $\Vit$ intersects $O$, and $\Vit$ provably contains all points in workspace reachable by the arm under the trajectory parameterized by $\ka$.
\end{proof}

The following theorem, the main result in this paper, confirms that feasible parameters for the constraints we generate are collision free and obey joint limits.
Note, we consider self-intersection below, in Appx. \ref{apdx:self_intersection}.

\begin{customthm}{\ref{thm:constraints_are_conservative}}
Any feasible solution to \eqref{prog:trajopt} parameterizes a trajectory that is collision-free and obeys joint limits over the time horizon $T$.
\end{customthm}
\begin{proof}
The conservatism of $h\obs$ follows from Lem. \ref{lem:zono_overapprox_rotatotope}, since each $\Zit$ is conservatively transformed into $\Vit$; $h\jlim$ is conservative by construction.
\end{proof}

\section{Safe Receding-Horizon Planning} \label{apdx:safe_planning}


ARMTD uses Alg. \ref{alg:online_planning} at each planning iteration.
Recall that, without loss of generality, each iteration generates a plan over the time horizon $T = [0,\tfin]$ (by shifting the current time to $0$).
Also recall, at each iteration, we allot $t\plan$ s within which to find a new plan.
The initial position and velocity of each joint in each iteration is the position and velocity, at time $t\plan$, of the trajectory plan of the previous iteration.
ARMTD attempts to find a safe trajectory within $t\plan$ by optimizing over a set of safe trajectory parameters; Thm. \ref{thm:constraints_are_conservative} ensures that any feasible solution is actually collision-free.
If no safe trajectory is found within the allotted time, the arm executes the braking maneuver specified by the previous safe trajectory.
Assuming the arm does not start in collision, this algorithm ensures that the arm is always safe (see \cite[Remark 70]{kousik2018_RTD_ijrr} or \cite[Theorem 1]{Hauser2012_receding_horizon}).

\section{Self-Intersection Constraints}\label{apdx:self_intersection}
Typical arms must avoid self-intersection between their links.
We specify $\selfidx \subset \N^2$ as a set of joint index pairs for which the links can intersect.
That is, for $(i,j) \in \selfidx$, there exist $q \in Q$ such that $\FO_i(q) \cap \FO_j(q) \neq \emptyset$.
\noindent For example, one may have $\selfidx = \{(1,3),(1,4),(2,4)\}$ for an arm with 4 links and three possible self-intersections.

We represent self-intersection constraints similarly to how we represent collision-avoidance constraints, with a function $h\self: \N\times\N\times T \times \Ka \to \R$.
Suppose $(i,j) \in \selfidx \subset \N^2$ indexes a pair of links that could intersect, whose volume is overapproximated by $\Vit$ and $\Vjt$.
In analogy to \eqref{eq:vit_split_slice_and_buff}, define
\begin{align}
    V\self(i,j,t) &= \Vitslc \oplus (-\Vjtslc) \quad \regtext{and} \\
    V\buf(i,j,t) &= \Vitbuf \oplus \Vjtbuf,
\end{align}
where $-\Vjtslc$ means the center and generators are multiplied by $-1$.
Let $A\self(i,j,t)$ and $b\self(i,j,t)$ return the half-space representation of $V\buf(i,j,t)$.
Then, using $*$ in place of the arguments $(i,j,t)$ for space,
\begin{align}
    h\self(*,\ka) = -\max\left(A\self(*)\eval(V\self(*),\ka) - b\self(*)\right).
\end{align}
As with $h\obs$, $h\self$ is a max of a linear combination of polynomials in $\ka$, so we can take the subgradient with respect to $\ka$.
Note one can prove a similar result to Lem. \ref{lem:hobs_is_conservative} for $h\self$.

With these self-intersection constraints, we again implement \eqref{prog:trajopt_general} as a nonlinear program, denoted \texttt{optTraj} in Alg. \ref{alg:online_planning}.
\begin{align}\label{prog:trajopt_self}
\begin{array}{cll}
    \underset{\ka \in \Ka}{\regtext{argmin}} & \costfunc(\ka) & \\
    \regtext{s.t.} & h\obs(i,t,O,\ka) < 0 & \forall\ i \in \{1,\cdots,n_q\},\ t \in T,\ O \in \obsset \\
    & h\self(i,j,t,\ka) < 0 & \forall\ (i,j) \in \selfidx,\ t \in T \\
    & h\jlim(\ka) < 0 & \forall\ i \in \{1,\cdots,n_q\}.
\end{array}
\end{align}

\section{Additional Explanations} \label{apdx:explanation}
\subsection{Forward Occupancy Example} \label{apdx:explanation:FO}
For an arm with $n_q > 2$, $\FO_i$ as in \ref{eq:forward_occupancy_i} can be written:
\begin{align}\begin{split}
    \FO_i(q)~=~&\Bigg\{R_1(q_1)l_1 + R_1(q_1)R_2(q_2)l_2 + \cdots\\
    &\cdots + \prod_{j\leq (i-1)}R_j(q_j)l_{i-1}\Bigg\}\oplus \left(\prod_{j\leq i}R_j(q_j)L_i\right).
\end{split}\end{align}
Notice that in this example, the rotated link volume of the $i$\ts{th} given by $\left(\prod_{j\leq i}R_j(q_j)L_i\right)$ is "stacked" on top of the sum of the positions of all predecessor joints.

\subsection{Slicing} \label{apdx:explanation:slicing}
ARMTD uses zonotopes and rotatotopes to represent RSs of parameterized trajectories of an arm.
In Sec. \ref{sec:online_planning}, our trajectory optimization implementation requires obtaining subsets of the RS corresponding to a particular choice of trajectory parameters.
We call this operation \defemph{slicing}, because it takes in a zono/rotatotope, evaluates some (or all) of its coefficients, and returns a zono/rotatotope that is a subset of the original with potentially fewer (or no) generators.

We define the $\slice$ function in Alg. \ref{alg:slice} using indeterminate evaluation and removal.
This function takes in a zono/rotatotope $Z = (x,g^i,\sym{\bt^i})^p$, a set of indeterminate coefficients $\{\sym{\sg^j}\}_{j=1}^m$, and a set of values for the indeterminate coefficients $\{\sg^j\}_{j=1}^m$, and outputs a sliced zono/rotatotope.
For each generator in $Z$, if $\sym{\sg^j}$ is a factor of that generator's coefficients (as in \eqref{eq:remove}), then the generator is multiplied by the value $\sg^j$.
If a generator becomes fully-sliced (and therefore has no more indeterminate coefficients), it is added to the center of the output zono/rotatotope, and removed from the set of generators.
For zonotopes, each generator becomes fully-sliced if its coefficient is evaluated because each coefficient has only one factor.
If a rotatotope is sliced until each generator has only one coefficient factor, the rotatotope becomes a zonotope.

To understand slicing, consider a particular choice of the trajectory parameter $\kvi$, as in \eqref{eq:slice_init_qdot}.
We want to obtain the subset of the JRS representing reachable joint angles corresponding to this particular trajectory parameter.
For each $\Zit$, only the generator $\givt$ is non-zero in the dimension corresponding to this trajectory parameter, meaning this $\kv$-sliceable generator is solely responsible for the volume of the reachable set in this dimension.
Choosing a particular value of the trajectory parameter means fixing this generator's indeterminate $\sym{\lkvit}$ to a particular value.
Since the $\kv$-sliceable generator is (generally) non-zero in the cosine and sine dimensions as well, the slicing operation returns a subset of the JRS in those dimensions (that is, by fixing the value of this indeterminate, we do not lose all of the JRS's volume in the cosine/sine dimensions, whereas the volume in the $k$-dimensions goes to zero).
For example, $\slice\left(\Zit,\left\{\sym{\lkvit}\right\},\left\{\frac{\pi}{6}\right\} \right)$ returns the subset of $\Zit$ corresponding to setting $\kvi = \frac{\pi}{6}$ rad/$s$ (provided that $\frac{\pi}{6} \in \Kvi$).

\subsection{Matrix Zonotope Example} \label{apdx:explanation:mat_zono}
This example constructs $\Mit$ from $\Zitv$ by reshaping the center and generators.
Suppose joint $i$ rotates about the $3$-axis of link $i-1$.
Then:, by \cite[(3.39)]{lavalle_textbook}, we have
\begin{align}\label{eq:mat_zono_construct_implementation}
    \Mit &= R_i(\initqi)\left(\Citv,\left\{\Gia(t),\Gjt\right\},\left\{\sym{\lkait},\sym{\btjt}\right\}\right)^{p(t)},\\
    \Citv &= \begin{bmatrix}
        c_i^{\regtext{v}} & -s_i^{\regtext{v}} & 0 \\
        s_i^{\regtext{v}} & c_i^{\regtext{v}} & 0 \\
        0 & 0 & 1
    \end{bmatrix} ,\ 
    \Gia(t) = \begin{bmatrix}
        c_i^{\regtext{a}} & -s_i^{\regtext{a}} & 0 \\
        s_i^{\regtext{a}} & c_i^{\regtext{a}} & 0 \\
        0 & 0 & 0 \\
    \end{bmatrix},\\
    \Gjt &= \begin{bmatrix}
        c_i^j & -s_i^j & 0 \\
        s_i^j & c_i^j & 0 \\
        0 & 0 & 0 \\
    \end{bmatrix}.
\end{align}
Since each $\Zitv$ is computed assuming $q_i(0 ;k) = 0$, we include $R_i(\initq_i)$ when constructing $\Mit$ to correct for each initial joint angle.
For different joint axes, $\Mit$ can be constructed accordingly \cite[Chapter 3.2.3]{lavalle_textbook}.

\subsection{Reduction of Generators} \label{apdx:explanation:reduction}
Creating the rotatotopes $\Vit$ in \eqref{eq:stacking} requires multiplying generators together and storing their product.
For example, a matrix zonotope described by $10$ matrices (a center and $9$ generators) multiplied by a zonotope described $2$ vectors (a center and $1$ generator) yields a rotatotope described by $20$ vectors (a center and $19$ generators).
Because this process is repeated for each joint, the number of generators theoretically required to represent each rotatotope grows exponentially with the number of joints.
In practice, many of these generators are very small, and their effect can be overapproximated without adding much conservatism to the RS.

We conservatively approximate \eqref{eq:stacking} by reducing the number of generators after each product, with a $\reduce$ function implemented as in \cite[Proposition 2.2 and Heuristic 2.1]{althoff2010reachability}.
The $\reduce$ function keeps the largest $n_\regtext{red}$ generators according to a user-defined metric (we used the $L^2$-norm), then overapproximates the rest of the generators with an axis-aligned box.
This ensures the number of rotatotope generators never exceeds a user-specified size.
From Lem. \ref{lem:one_k_sliceable_gen_per_tope}, each $\Mit$ has $\ka$-sliceable generators.
If a $\ka$-sliceable generator is chosen for reduction, we no longer consider it $\ka$-sliceable.
This is a conservative approach, because slicing reduces the volume of a rotatotope in Alg. \ref{alg:slice}.
A generator that is no longer $\ka$-sliceable cannot decrease the volume of the RS for any choice of $\ka$.



\subsection{Hard Scenarios} \label{apdx:explanation:hard_scenarios}
\begin{figure*}[!ht]
    \centering
    \includegraphics[width=\textwidth]{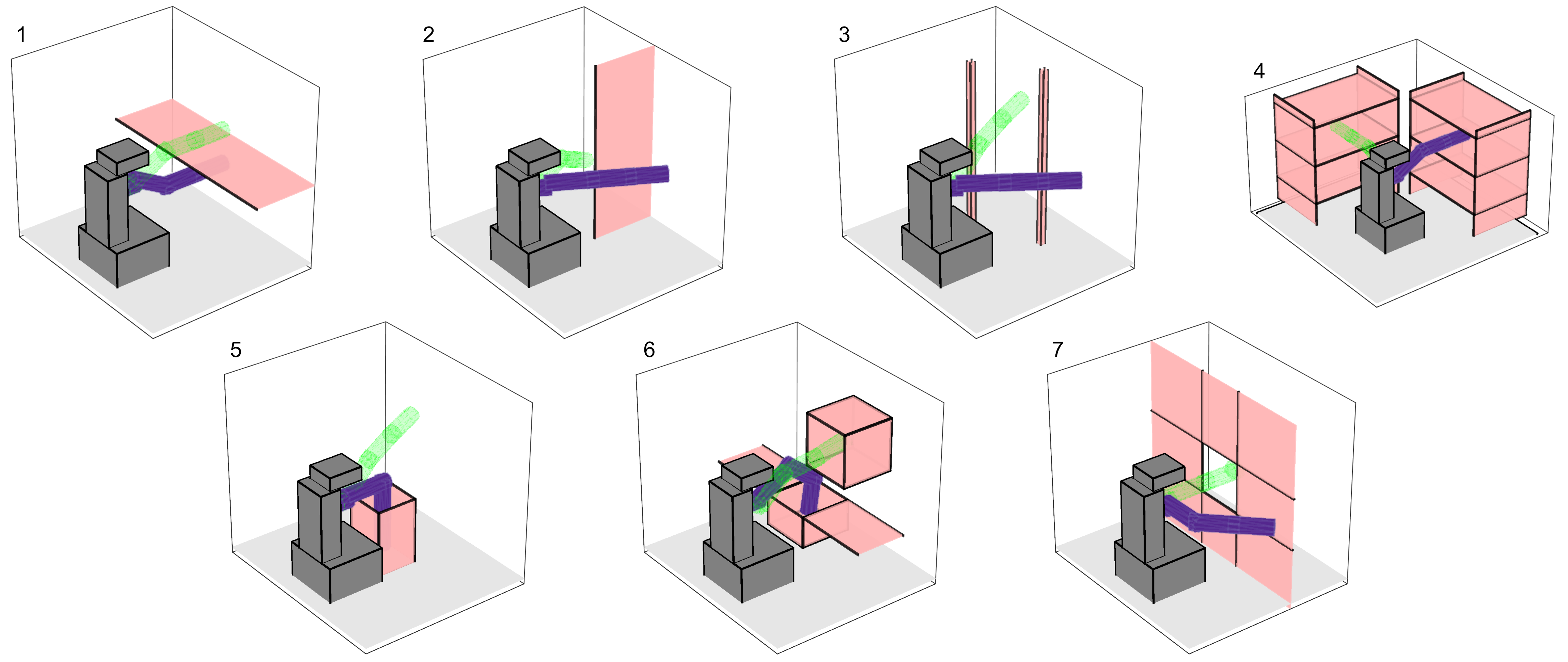}
    \caption{The set of seven Hard Scenarios (number in the top left), with start pose shown in purple and goal pose shown in green.
    There are seven tasks in the Hard Scenarios set: (1) from below to above a table, (2) from one side of a wall to another, (3) between two vertical posts, (4) from one set of shelves to another, (5) from inside to outside of a box on the ground, (6) from a sink to a cupboard, (7) through a small window.}
    \label{fig:hard_scenarios}
\end{figure*}

The set of Hard Scenarios is shown in Fig. \ref{fig:hard_scenarios}.

\subsection{Design Choices and Hyperparameters} \label{apdx:explanation:design}
ARMTD has several design choices and hyperparameters, all of which can impact the time required for online planning, but none of which impact the strict safety guarantees.
That is, \textbf{ARMTD guarantees safety independent of design choices}.

The first design choice to consider is the trajectory parameterization.
While we provide a generic definition (Def. \ref{def:traj_param_generic}), we find that parameterizing velocities and accelerations in our implementation provides a physical intuition for the planned trajectories.
For future work, we plan to explore other parameterizations that provide, for example, smoother motion profiles.

The next design choice to consider are those that define the user-specified cost function, generated at each receding-horizon planning iteration.
A more non-convex cost function can slow down online planning.
In this work, we generate the cost function by using a high-level planner (HLP) such as an RRT* to generate waypoints between the robot's current location and the global goal.
Importantly, the waypoints need not be collision-free; they are used to create a cost function that rewards reaching the waypoint, but ARMTD's safety constraints take care of collision-avoidance.
Therefore, \textbf{ARMTD provides a safety layer on top of RRT*} or any other HLP (e.g., PRM, or simply picking a waypoint along a straight-line between the robot and the goal).

There are two hyperparameters that determine a tradeoff between conservatism and online planning speed (without impacting safety).
The first is the density of the time partition for the JRS.
That is, if we partition time more finely to generate the zonotope JRS, then it takes longer to generate and evaluate constraints at runtime (because we have to consider more zonotopes), but the JRS is also less conservative (so, the robot has more free space to move through).

The second hyperparameter is the range of parameters in the trajectory parameterization.
A larger range produces a more conservative JRS, because the same number of zonotopes (determined by the time partition) must contain a larger range of joint angles achieved by all parameterized trajectories.
We mitigate this problem in practice by precomputing many JRSs (in this work, we used $400$), each of which has a narrow range of initial velocity parameters $\Kvi$.
We choose the range of acceleration parameters $\Kai$ to vary with the velocity parameters, so that at higher speeds, there is a larger range of available control actions.
This reduces conservatism at lower speeds so that ARMTD can maneuver tightly around obstacles.

Note, each JRS only takes around $1$ s to compute, since it is only for a single joint, and for the low-dimensional cosine/sine dynamics.
At runtime, to construct the RS of the entire arm, we first select the JRS (for each joint) containing the current initial velocity within its narrow range.
Then, we slice by the exact initial velocity to produce the RS, and the corresponding collision-avoidance constraints.

\subsection{Seeding CHOMP with RRT*} \label{apdx:explanation:seeding_chomp}
CHOMP performs better when seeded with a path output by RRT*, as opposed to the default straight-line initialization \cite{chomp,moveit}.
Given that ARMTD uses RRT* to generate waypoints at each receding-horizon planning iteration, one may wonder why we do not use the same RRT* to seed CHOMP.
However, ARMTD and CHOMP use RRT* in fundamentally different ways.
ARMTD plans in a receding-horizon way, so its runtime and safety guarantees are not dependent on the RRT* output.
On the other hand, CHOMP would require the RRT* to run for some (unknown) duration, then perform trajectory optimization.
In other words, CHOMP requires additional planning time for seeding, whereas ARMTD does not.
So, in terms of the most important metric in this work (finding a collision-free trajectory in under $t\plan = 0.5$ s), it is unclear how much time to dedicate for RRT* and how much for CHOMP.

The challenge of generating a fair comparison is compounded by the fact that ARMTD does not require the output of the RRT* to be collision-free.
One could potentially use CHOMP in a receding-horizon way, by attempting to reach an intermediate waypoint generated by RRT* in each planning iteration.
But, the available open-source CHOMP implementation (via MoveIt! \cite{moveit}) requires the goal (i.e., intermediate waypoint) to be collision-free.
Implementing CHOMP in a more generalized receding-horizon framework is outside the scope of the present work.